\documentclass[twoside,11pt]{article}

\usepackage{pdflscape}
\usepackage{booktabs}
\usepackage{multirow}
\usepackage{paralist,amsmath, amssymb, bm}
\usepackage{algorithm,algorithmic}
\usepackage{jmlr2e}

\usepackage[colorlinks,
            linkcolor=blue,
            citecolor=blue,
            urlcolor=magenta,
            linktocpage,
            plainpages=false]{hyperref}

\makeatletter
\newcounter{ALC@tempcntr}
\makeatother

\makeatletter
\AtBeginDocument{\Hy@breaklinkstrue}
\makeatother

\newtheorem{ass}{Assumption}

\def \x {\mathbf{x}}

\def \u {\mathbf{u}}

\def \w {\mathbf{w}}
\def \bv {\mathbf{v}}
\def \R {\mathbb{R}}

\def \W {\mathcal{W}}

\def \A {\mathcal{A}}

\def \bv {\mathbf{v}}

\def \I {\mathcal{I}}

\def \B {\mathbf{B}}
\def \tb {\tilde{b}}

\def \tg {\tilde{g}}

\def \bb {\mu}

\def \B {\mathcal{B}}

\def \ind {\mathbb{I}}

\def \leh {n}

\DeclareMathOperator*{\sgn}{sign}
\DeclareMathOperator*{\erf}{erf}

\DeclareMathOperator*{\RS}{R-S}
\DeclareMathOperator*{\ARS}{A-R-S}

\begin{document}

\title{Smoothed Online Convex Optimization Based on Discounted-Normal-Predictor}

\author{\name Lijun Zhang \email zhanglj@lamda.nju.edu.cn\\
        \name Wei Jiang \email jiangw@lamda.nju.edu.cn\\
       \addr National Key Laboratory for Novel Software Technology, Nanjing University, Nanjing 210023, China\\
        \name Jinfeng Yi  \email yijinfeng@jd.com \\
      \addr JD.com, Beijing 100101, China \\
       \name Tianbao Yang \email tianbao-yang@uiowa.edu\\
       \addr Department of Computer Science, the University of Iowa, Iowa City, IA 52242, USA
}
\editor{}

\maketitle

\begin{abstract}
In this paper, we investigate an online prediction strategy named as Discounted-Normal-Predictor \citep{DNP:2010:Arxiv} for smoothed online convex optimization (SOCO), in which the learner needs to minimize not only the hitting cost but also the switching cost. In the setting of learning with expert advice, \citet{pmlr-v98-daniely19a} demonstrate that Discounted-Normal-Predictor can be utilized to yield nearly optimal regret bounds over any interval, even in the presence of switching costs. Inspired by their results, we develop a simple algorithm for SOCO: \emph{Combining online gradient descent (OGD) with different step sizes sequentially by Discounted-Normal-Predictor}. Despite its simplicity, we prove that it is able to minimize the adaptive regret with switching cost, i.e., attaining nearly optimal regret with switching cost on every interval. By exploiting the theoretical guarantee of OGD for dynamic regret, we further show that the proposed algorithm can minimize the dynamic regret with switching cost in every interval.
\end{abstract}

\section{Introduction}
Recently, a variant of online convex optimization (OCO), named as smoothed OCO (SOCO) has received lots of attention in the machine learning community \citep{NIPS2019_8463,NEURIPS2020_a6e4f250}. In each round $t$, the online learner chooses an action $\w_t$ from a convex domain $\W$, and a convex loss function $f_t(\cdot):\W \mapsto \R$ is selected by an adversary. While in the traditional OCO, the learner only suffers a \emph{hitting} cost $f_t(\w_t)$, in SOCO, it further pays a \emph{switching} cost $m(\w_t,\w_{t-1})$, which penalizes the learner for changing its actions between rounds. SOCO has found wide applications in real-world problems where the change of states usually brings additional costs \citep{Geographical:Load:Balance,electricity:pricing,Spatiotemporal:Prediction,Optimization:timescales}, such as the wear-and-tear cost of switching servers \citep{Dynamic:Data:Center}.

For general convex functions, a natural choice of the switching cost is the distance between successive actions measured by the $\ell_2$-norm, i.e., $m(\w_t,\w_{t-1})=\|\w_t-\w_{t-1}\|$. Then, the total loss in the $t$-th round becomes
\begin{equation} \label{eqn:total:loss}
f_t(\w_t) + \lambda G \|\w_t-\w_{t-1}\|
\end{equation}
where $\lambda \geq 0$ is the trade-off parameter, and $G$ is the upper bound of the norm of gradients. Here, $G$ is introduced to ensure that the two terms weighted by $\lambda$ are on the same order.  Following the convention of online learning \citep{bianchi-2006-prediction}, we choose regret to measure the performance, and meanwhile take into account the switching cost. Let $T$ be the total number of iterations. The standard metric would be \emph{regret with switching cost}:
\begin{equation} \label{eqn:reg:switch}
\sum_{t=1}^T \big( f_t(\w_t) + \lambda G \|\w_t-\w_{t-1}\| \big) - \min_{\w \in \W} \sum_{t=1}^T f_t(\w)
\end{equation}
defined as the difference between the cumulative loss of the learner and that of the best fixed decision in hindsight. It is easy to verify that online gradient descent (OGD) \citep{zinkevich-2003-online} attains an $O(\sqrt{(1+\lambda)T})$ bound for regret with switching cost (c.f.~Theorem~\ref{thm:regret:ogd}), which can be proved to be optimal \citep[Theorem 4]{arXiv:2007.04393}.

However, regret is not suitable for changing environments in which the best decision may change over time. To address this limitation, new performance measures, including adaptive regret and dynamic regret have been proposed \citep{IJCAI:2020:Zhang,Online:Review:Casa}. Adaptive regret enforces the algorithm to have a small regret over every interval \citep{Adaptive:Hazan,Adaptive:ICML:15}, which essentially measures the performance w.r.t.~a changing comparator. By extending adaptive regret to SOCO, we obtain \emph{adaptive regret with switching cost}:
\begin{equation} \label{eqn:adaptive:regret:switch}
\ARS(T, \tau)=\max_{r \leq T+1-\tau}  \RS(r,r+\tau-1)
\end{equation}
where $\tau$ is the length of the interval, and
\[
\RS(r,s)= \sum_{t=r}^{s} \big( f_t(\w_t) + \lambda G \|\w_t-\w_{t-1}\| \big) -  \min_{\w \in \W}  \sum_{t=r}^{s}  f_t(\w)
\]
is the regret with switching cost over interval $[r,s]$. To cope with changing environments, dynamic regret directly compares the learner against a sequence of comparators $\u_1, \ldots,\u_T \in \W$ \citep{zinkevich-2003-online}. Similarly, we can incorporate the switching cost into dynamic regret, and obtain \emph{dynamic regret with switching cost}:
\begin{equation} \label{eqn:dynamic:regret:switch}
\sum_{t=1}^T \big( f_t(\w_t) + \lambda G \|\w_t-\w_{t-1}\| \big) - \sum_{t=1}^T f_t(\u_t).
\end{equation}

In the literature, there are only two works that investigate adaptive regret with switching cost. However, one of them relies on strong convexity \citep{arXiv:2007.04393}, and the other one makes use of sophisticated techniques \citep{arXiv:2102.01623}. There also exist preliminary studies on dynamic regret with switching cost, but are limited to the special case that $\lambda=1/G$ \citep{Smoothed:Online:NeurIPS} or follow different settings \citep{Prediction:Switching,NIPS2019_8463}. Besides, all of previous works only target one performance measure. Notice that for OCO, we do have algorithms that are able to minimize adaptive regret and dynamic regret simultaneously \citep{Adaptive:Dynamic:AISTATS,pmlr-v119-cutkosky20a}.

In this paper, we develop a simple algorithm for SOCO based on Discounted-Normal-Predictor \citep{DNP:2010:Arxiv,NIPS2011_11b921ef}, which is designed for online bit prediction and can be used to combine two experts. In a recent study, \citet{pmlr-v98-daniely19a} reveal that Discounted-Normal-Predictor automatically controls the switching cost, and with the help of projection, can be extended to support adaptive regret. Motivated by their observations, we first analyze a variant of Discounted-Normal-Predictor with conservative updating (abbr.~DNP-cu), and prove that it suffers a small loss on every interval, even in the presence of switching costs. To ensure adaptivity, we choose conservative updating instead of projection because the former one can be analyzed more easily.  Then, we create multiple OGD with different step sizes, and combine them sequentially by DNP-cu. Theoretical analysis shows that the proposed method achieves an $O(\sqrt{(1+\lambda) \tau \log T })$ bound for adaptive regret with switching cost. Furthermore, based on the fact that OGD is also equipped with dynamic regret bounds, we establish nearly optimal guarantees for dynamic regret with switching cost in every interval. Specifically, for any interval $[r,s]$ with length $\tau$, our method attains an $O(\sqrt{(1+\lambda) \tau (1+P_{r,s})\log T })$  bound for dynamic regret with switching cost, 
where
\begin{equation} \label{eqn:path:length:interval}
P_{r,s}= \sum_{t=r}^s \|\u_t-\u_{t+1}\|
\end{equation}
is the path-length of  an arbitrary comparator sequence $\u_r,\ldots,\u_s\in \W$.

Compared with state-of-the-art results, this paper has the following advantages.
\begin{compactenum}
\item Our $O(\sqrt{(1+\lambda) \tau \log T })$ adaptive regret bound is on the same order as that of \citet{arXiv:2102.01623}, but our method is more simple. In contrast, their algorithm pieces together several techniques, including online coin betting, geometric covering intervals, and the reduction from unconstrained online learning to constrained online learning.
\item Although \citet{Smoothed:Online:NeurIPS} establish an $O(\sqrt{T (1+P_{1,T}) })$ dynamic regret bound with switching cost, they only consider the case that $\lambda=1/G$ and the whole interval $[1,T]$. By comparison, our dynamic regret bound holds for any $\lambda\geq 0$ and any interval.
\item To the best of our knowledge, this is the first effort to minimize both adaptive regret and dynamic regret, under the setting of SOCO.
\end{compactenum}

Finally, we would like to emphasize the strength of Discounted-Normal-Predictor, which, in our opinion, was not getting enough attention. The results of \citet{pmlr-v98-daniely19a} and this paper demonstrate that Discounted-Normal-Predictor provides an elegant way (and a very different methodology) to minimize both adaptive regret and dynamic regret, with or without switching cost, under the setting of learning with expert advice (LEA) or OCO.

\section{Related Work}
We briefly review the related work on OCO and SOCO, as well as Discounted-Normal-Predictor.
\subsection{Online Convex Optimization (OCO)}
During the past decades, OCO has been extensively studied, and various algorithms have been proposed to minimize the regret, i.e., (\ref{eqn:reg:switch}) with $\lambda=0$ \citep{Online:suvery,Intro:Online:Convex}. It is well-known that OGD \citep{zinkevich-2003-online} achieves an $O(\sqrt{T})$ bound which is minimax optimal \citep{Minimax:Online}. We can obtain tighter regret bounds if the loss function satisfies specifical curvature properties, such as strong convexity \citep{ICML_Pegasos}, exponential concavity  \citep{ML:Hazan:2007,Beyond:Logarithmic}, and smoothness \citep{NIPS2010_Smooth,Gradual:COLT:12}.

Adaptive regret has been explored under the setting of LEA \citep{LITTLESTONE1994212,Freund:1997:UCP,Adamskiy2012,Track_Large_Expert,pmlr-v40-Luo15} and OCO \citep{Hazan:2009:ELA,jun2017,Adaptive:Regret:Smooth:ICML,Dual:Adaptivity}. The seminal work of \citet{Adaptive:Hazan} firstly introduces adaptive regret to OCO, and defines it as the maximum regret over all intervals:
\begin{equation} \label{eqn:adaptive:regret}
\max_{[r,s] \subseteq [T]} \left(\sum_{t=r}^{s} f_t(\w_t) -  \min_{\w \in \W}  \sum_{t=r}^{s}  f_t(\w) \right).
\end{equation}
However, (\ref{eqn:adaptive:regret}) is dominated by long intervals, and does not respect short intervals well. To avoid this issue, \citet{Adaptive:ICML:15} propose a refined definition, which takes the interval length $\tau$ as a parameter. When $\lambda=0$, (\ref{eqn:adaptive:regret:switch}) reduces to their formulation. For general convex functions, \citet{Improved:Strongly:Adaptive} establish an  $O(\sqrt{\tau \log T})$  bound, which holds for any interval length $\tau$.

Dynamic regret is proposed by \citet{zinkevich-2003-online}, and has a similar spirit with the tracking regret in LEA \citep{LITTLESTONE1994212,Herbster1998,Fixed:Share:NIPS12}. Its definition can be found by setting $\lambda=0$ in (\ref{eqn:dynamic:regret:switch}). Existing works have analyzed two types of dynamic regret: (i) the general version where the comparator sequence $\u_1, \ldots, \u_T$ is arbitrary \citep{Dynamic:ICML:13,Adaptive:Dynamic:Regret:NIPS,Problem:Dynamic:Regret,pmlr-v134-baby21a}, and (ii) the worst-case version where $\u_t$ is chosen as the minimizer of $f_t(\cdot)$ \citep{Dynamic:AISTATS:15,Non-Stationary,Dynamic:2016,Dynamic:Strongly,Dynamic:Regret:Squared,Worst:Dynamic:SSS}. This paper will focus on the general dynamic regret, since it includes the standard regret and the worst-case dynamic regret as special cases. For general convex functions, Ader attains an optimal $O(\sqrt{T(1+P_{1,T})})$ bound, where $P_{1,T}$ is the path-length of $\u_1,\ldots,\u_T$ \citep{Adaptive:Dynamic:Regret:NIPS}.

Both adaptive regret and dynamic regret are designed for changing environments, but our understanding of their relationship is quite limited. Currently, we know that it is possible to bound the worst-case dynamic regret by adaptive regret \citep{Dynamic:Regret:Adaptive}, and minimize  adaptive regret and the general dynamic regret simultaneously \citep{Adaptive:Dynamic:AISTATS,pmlr-v119-cutkosky20a}.
\subsection{Smoothed Online Convex Optimization (SOCO)}
The research of SOCO is motivated by real-world applications where the switching cost plays a crucial role \citep{Dynamic:Data:Center,Geographical:Load:Balance}. Besides regret, competitive ratio is another popular metric for SOCO, and a large number of algorithms have been proposed to yield dimension-free competitive ratio. However, all of them are limited to the \emph{lookahead} setting where the learner can observe the hitting cost $f_t(\cdot)$ before deciding its action $\w_t$. Furthermore, their analysis relies  on strong conditions, including low dimensionality \citep{bansal_et_al:LIPIcs,10.1007/978-3-319-89441-6_13},  polyhedrality \citep{SOCO:OBD}, quadratic growth \citep{NIPS2019_8463} and strong convexity \citep{OBD:Strongly,pmlr-v125-argue20a}, and fail on general convex functions. We note that SOCO is closely related to convex body chasing (CBC) \citep{Convex:Body,CCBF:2016,NCB:2018,Linear:CNCB,CCCB:Bubeck,Optimal:CNCB,Linear:CCB,Optimal:CCB}. For more details, please refer to  \citet{10.1145/3379484}.

In the study of online control, \citet{arXiv:2007.04393} and \citet{arXiv:2102.01623} have developed adaptive algorithms for OCO with memory, which can be applied to bounding adaptive regret with switching cost. However, \citet{arXiv:2007.04393} assume the hitting cost is strongly convex, and follow the definition in (\ref{eqn:adaptive:regret}). \citet{arXiv:2102.01623} successfully deliver an $O(\sqrt{(1+\lambda) \tau \log T })$ bound for adaptive regret with switching cost, but their method is rather complex. Specifically, they make use of online coin betting \citep{NIPS2016_32072254}, geometric covering intervals \citep{Adaptive:ICML:15}, and the reduction from unconstrained online linear optimization to constrained online linear optimization \citep{pmlr-v75-cutkosky18a}.

For dynamic regret with switching cost, \citet{Smoothed:Online:NeurIPS} extend Ader to support switching costs, and prove an optimal  $O(\sqrt{T(1+P_{1,T})})$ bound, but they only consider the case that $\lambda=1/G$. Other works on dynamic regret with switching cost are incomparable to our paper, because they either rely on strong convexity \citep{Prediction:Switching,NEURIPS2020_a6e4f250},  set the switching cost as the squared distance \citep{NIPS2019_8463}, or assume an upper bound of the total switching cost is given in advance \citep{SOCO:OBD,10.1145/3375788}.
\begin{algorithm}[t]
\caption{Discounted-Normal-Predictor}
\begin{algorithmic}[1]
\REQUIRE Two parameters: $\leh$ and $Z$
\STATE Set $x_1=0$, and $\rho=1-1/\leh$
\FOR{$t=1,\ldots,T$}
\STATE Predict $g(x_t)$
\STATE Receive $b_t$
\STATE Set $x_{t+1}=\rho x_t + b_t$
\ENDFOR
\end{algorithmic}
\label{alg:1}
\end{algorithm}
\subsection{Discounted-Normal-Predictor}
Following the terminology of \citet{DNP:2010:Arxiv}, we introduce Discounted-Normal-Predictor in the context of the bit prediction problem. Let $b_1,\ldots,b_T$ be an adversarial sequence of bits, where $b_t \in [-1,1]$ can take real values. In each round $t$, the algorithm is required to output a confidence level $c_t \in [-1,1]$,  then observes the value of $b_t$, and finally gets a \emph{payoff} $c_t b_t$. The goal is to maximize the cumulative payoff of the algorithm $\sum_{t=1}^T c_t b_t$.

Let $\leh >0$ be a parameter for the interval length, and define the discount factor
\begin{equation} \label{eqn:discount}
\rho=1-\frac{1}{\leh}.
\end{equation}
As the name suggests, Discounted-Normal-Predictor maintains a discounted deviation
\[
x_t = \sum_{j=1}^{t-1} \rho^{t-1-j} b_j
\]
at each round $t$, and the prediction is determined by $g(x_t)$ for a confidence function $g(\cdot)$ defined as
\begin{equation} \label{eqn:confidence:original}
g(x)=\sgn(x) \cdot \min \left( Z \cdot \erf\left( \frac{|x|}{4 \sqrt{\leh}} \right) e^{\frac{x^2}{16 \leh}} , 1\right)
\end{equation}
where  $Z>0$ is a parameter, and $\erf(x) = \frac{2}{\sqrt{\pi}} \int_0^x e^{-s^2} d s$ is the error function. The complete procedure is summarized in Algorithm~\ref{alg:1}.

For any $Z \leq 1/e$, \citet[Theorem 14]{DNP:2010:Arxiv} have proved that Discounted-Normal-Predictor satisfies
\[
\begin{split}
\sum_{t=1}^T g(x_t) b_t  \geq \max\left( \left|\sum_{j=1}^T b_j \right| -O \big(\sqrt{T \log (1/Z)}\big), - O\big(Z\sqrt{T}\big)\right)
\end{split}
\]
where we set $\leh=T$ in Algorithm~\ref{alg:1}. By choosing $Z=o(1/T)$, we observe that it has $O(\sqrt{T \log T})$ regret against the strategy that predicts the majority bit (whose payoff is $|\sum_{j} b_j |$), as well as a subconstant $o(1)$ loss. 
To address the problem of learning with two experts, we can define $b_t$ as the difference between the losses of experts and restrict $c_t \in [0,1]$. It follows that the algorithm suffers $O(\sqrt{T \log T})$ and $o(1)$ regret w.r.t.~the two experts, respectively \citep[Lemma 15]{DNP:2010:Arxiv}. Discounted-Normal-Predictor can be applied to the general setting of $N$ experts, by aggregating experts one by one. We note that the problem of trading off regret to the best expert for regret to the ``special'' expert  stems from the study of \citet{COLT:07:Average}, and is later investigated by \citet{NIPS2014_Easy}.

\citet[Theorem 5]{DNP:2010:Arxiv} also investigate a variant of the adaptive regret, which uses an infinite window with geometrically decreasing weighting. To this end, they propose a conservative updating rule to control the value of the deviation $x_t$. The current bit $b_t$ is utilized to update $x_t$ only when the confidence of the algorithm is low, or when the algorithm predicts incorrectly. To be specific, Line 5 of Algorithm~\ref{alg:1} is replaced by
\begin{algorithmic}
\IF{$|x_t| < U(\leh)$ or $g(x_t)b_t <0$}
\STATE Set $x_{t+1}=\rho x_t + b_t$
\ELSE
\STATE Set $x_{t+1}=\rho x_t$
\ENDIF
\end{algorithmic}
where $U(\leh)=O(\sqrt{\leh \log (1/Z)})$ is a constant such that $g(x)=1$ for $|x| \geq U(\leh)$. However, it remains open whether their conservative updating works with the standard adaptive regret \citep{Adaptive:ICML:15}, which is answered affirmatively by our Theorem~\ref{thm:1}.


\citet{pmlr-v98-daniely19a} extend Discounted-Normal-Predictor to support the switching cost and adaptive regret. The confidence function is modified slightly as\footnote{Their definition of the error function drops the constant $\frac{2}{\sqrt{\pi}}$. }
\begin{equation} \label{eqn:confidence:new}
g(x)=\Pi_{[0,1]} \left[\tg(x) \right]
\end{equation}
where
\begin{equation} \label{eqn:confidence:new2}
\tg(x)=\sqrt{\frac{\leh}{8}} Z \cdot \erf\left( \frac{x}{\sqrt{8 \leh}} \right) e^{\frac{x^2}{16 \leh}},
\end{equation}
and $\Pi_{[0,1]} [ \cdot ]$ denotes the projection operation onto the set $[0,1]$. First, by a more careful analysis, they demonstrate that Discounted-Normal-Predictor has similar regret bounds even in the presence of switching costs. Second, they introduce a projection operation to prevent $x_t$ from being too large, and then derive tight bounds for the standard adaptive regret. Specifically, the updating rule for $x_t$ becomes
\[
x_{t+1}=\Pi_{[-2, U(\leh)+2]} \left[\rho x_t + b_t \right]
\]
where
\begin{equation} \label{eqn:u:tau}
U(\leh)=\tg^{-1}(1) \leq \sqrt{16\leh \log \frac{1}{Z}}.
\end{equation}
Our work is inspired by \citet{pmlr-v98-daniely19a}, but with the following differences.
\begin{compactitem}
\item While \citet{pmlr-v98-daniely19a} consider the setting of LEA, we investigate OCO.
\item To bound the adaptive regret, \citet{pmlr-v98-daniely19a} introduce the projection operation. In contrast, we make use of the conservative updating of \citet{DNP:2010:Arxiv}, whose effect can be analyzed more easily.
\item We study not only the adaptive regret, but also the dynamic regret. Our algorithm is equipped with nearly optimal bounds for both metrics, in the presence of switching costs.
\end{compactitem}

\section{Main Results}
We take Discounted-Normal-Predictor with conservative updating (DNP-cu) as our meta-algorithm, and analyze its behavior by taking the switching cost into consideration. Then, we use DNP-cu to combine multiple OGD sequentially, and present its adaptive regret and dynamic regret with switching cost.

\subsection{The Meta-algorithm}
To consist with previous studies, we describe DNP-cu from the perspective of bit prediction, but require the prediction to lie in $[0,1]$ so that it can be used later as a meta-algorithm to combine experts. The detailed procedure is summarized in Algorithm~\ref{alg:2}. Compared with the original algorithm \citep{DNP:2010:Arxiv}, we make two slight modifications.
\begin{compactenum}
\item We choose the confidence function $g(\cdot)$ in (\ref{eqn:confidence:new}), whose property has been revealed by \citet{pmlr-v98-daniely19a} more formally.
\item We adapt the updating rule to the fact that $g(\cdot)$ belongs to $[0,1]$ instead of $[-1,1]$. Specifically, we perform the standard updating in Line 6, when the confidence is low i.e., $x_t \in [0, U(\leh)]$, or when the algorithm predicts incorrectly i.e., 
    \[
    x_t <0 \cap  b_t > 0 \textrm{ or } x_t >U(\leh) \cap b_t < 0.
    \]
     Otherwise, we follow Line 8, which only shrinks $\x_t$ and ignores $b_t$.
\end{compactenum}

\begin{algorithm}[t]
\caption{Discounted-Normal-Predictor with conservative updating (DNP-cu)}
\begin{algorithmic}[1]
\REQUIRE Two parameters: $\leh$ and $Z$
\STATE Set $x_1=0$, $\rho=1-1/\leh$, and $U(\leh)=\tg^{-1}(1)$
\FOR{$t=1,\ldots,T$}
\STATE Predict $g(x_t)$ where $g(\cdot)$ is defined in (\ref{eqn:confidence:new})
\STATE Receive $b_t$
\IF{$x_t \in [0, U(\leh)]$ or $x_t <0 \cap  b_t > 0$ or $x_t >U(\leh) \cap b_t < 0$}
\STATE Set $x_{t+1}=\rho x_t + b_t$
\ELSE
\STATE Set $x_{t+1}=\rho x_t$
\ENDIF
\ENDFOR
\end{algorithmic}
\label{alg:2}
\end{algorithm}
To analyze the performance of DNP-cu, we consider both the payoff and the switching cost, referred to as \emph{reward} below. We have the following theorem which exploits the possibility that the magnitude of the bit sequence may be smaller than $1$.
\begin{theorem} \label{thm:1} Suppose $Z \leq \frac{1}{e}$ and $\leh \geq \max\{8e, 16\log \frac{1}{Z}\}$. For any bit sequence $b_1,\ldots,b_T$ such that $|b_t| \leq \mu \leq 1$, the cumulative reward of Algorithm~\ref{alg:2} over any interval $[r,s]$ with length $\tau$ satisfies
\begin{equation} \label{eqn:alg2:lower}
\begin{split}
&\sum_{t=r}^s \left( g(x_t) b_t -   \frac{1}{\bb} |g(x_t)-g(x_{t+1})| \right)\\
\geq & \max\left(0, \sum_{t=r}^s b_t   - \frac{\tau}{\leh} \left( U(\leh) +2 \mu\right) - U(\leh) -\mu \right) -  U(\leh) -\mu - Z \tau
\end{split}
\end{equation}
where $U(\leh)$ is defined in (\ref{eqn:u:tau}). Furthermore, for intervals starting from $1$, we have
\begin{equation} \label{eqn:alg2:lower:second}
\begin{split}
&\sum_{t=1}^s \left(g(x_t) b_t -   \frac{1}{\bb} |g(x_t)-g(x_{t+1})| \right)\\
\geq & \max\left(0, \sum_{t=1}^s b_t   - \frac{\tau(U(\leh) +2 \mu)}{\leh}   - U(\leh) \right)- Z \tau .
\end{split}
\end{equation} And the change of successive predictions satisfies
\begin{equation} \label{eqn:change:prediction:F:2}
|g(x_t)-g(x_{t+1})| \leq  \bb  \sqrt{\frac{1}{\leh} \log \frac{1}{Z}}+ \frac{Z \bb}{4}.
\end{equation}
\end{theorem}
\textbf{Remark: } First, the above theorem reveals that DNP-cu controls the switching cost automatically, and the trade-off between the payoff and the switching cost is determined by the magnitude of the bit sequence. Second, thanks to the conservative updating, we are able to bound the cumulative reward over any interval, which can be exploited to support adaptive regret.

Recall that in the total loss (\ref{eqn:total:loss}), the trade-off parameter $\lambda$ could be arbitrary. As a result, to utilize DNP-cu for SOCO, we also need a flexible way to balance the payoff and the switching cost. To this end, we introduce $\lambda$ and define the reward in the $t$-round as
\begin{equation}\label{eqn:reward:t}
g(x_t) b_t - \lambda |g(x_t)-g(x_{t-1})|.
\end{equation}
Then, our goal is to maximize
\begin{equation}\label{eqn:reward:interval}
\sum_{t=r}^s \big( g(x_t) b_t -   \lambda |g(x_t)-g(x_{t-1})| \big)
\end{equation}
for each interval $[r,s]$. Based on Theorem~\ref{thm:1},  a straightforward way is to multiply the bit sequence by $1/\lambda$, and then pass it to Algorithm~\ref{alg:2}. But in this way, the lower bound will scale linearly with $\lambda$. To improve the dependence on $\lambda$, we will multiply the bit sequence by $1/\sqrt{\lambda}$ \citep{pmlr-v98-daniely19a}. Specifically, we have the following corollary based on (\ref{eqn:alg2:lower}) and (\ref{eqn:change:prediction:F:2}) of Theorem~\ref{thm:1}.\footnote{We omit (\ref{eqn:alg2:lower:second}), because it is not used in the subsequent analysis.}
\begin{corollary} \label{cor:1} Suppose $Z \leq \frac{1}{e}$, $\leh \geq \max\{8e, 16\log \frac{1}{Z}\}$, and $b_1,\ldots,b_T$ is a bit sequence such that $|b_t| \leq 1$. Running Algorithm~\ref{alg:2} over the \emph{scaled} bit sequence
\[
\frac{b_1}{\max(\sqrt{\lambda},1)},\ldots, \frac{b_T}{\max(\sqrt{\lambda},1)},
\]
for any interval $[r,s]$ with length $\tau$, we have
\begin{equation} \label{eqn:cor:lower}
\begin{split}
&\sum_{t=r}^s \big( g(x_t) b_t -   \lambda |g(x_t)-g(x_{t+1})| \big)\\
\geq & \max\left(0, \sum_{t=r}^s b_t   -  \frac{\max(\sqrt{\lambda},1) U(\leh)(\tau+\leh)}{\leh}  - \frac{2\tau+\leh}{\leh}   \right) \\
& - \max(\sqrt{\lambda},1) U(\leh) -1  -\max(\sqrt{\lambda},1) Z \tau.
\end{split}
\end{equation}
And the change of successive predictions satisfies
\begin{equation} \label{eqn:change:prediction:F:3}
\begin{split}
 |g(x_t)-g(x_{t+1})| \leq  \frac{1}{\max(\sqrt{\lambda},1)}  \left( \sqrt{\frac{1}{\leh} \log \frac{1}{Z}}+ \frac{Z}{4} \right) .
\end{split}
\end{equation}
\end{corollary}
\textbf{Remark: } First, there is a slight difference between (\ref{eqn:reward:interval}) and the reward in the 1st line of (\ref{eqn:cor:lower}). For brevity, we do not distinguish between them, because they only differ by a constant factor. Second, we discuss the implications of (\ref{eqn:cor:lower}).  From the definition of $U(\leh)$ in (\ref{eqn:u:tau}), we  have
\begin{equation}\label{eqn:al2:prop1}
\begin{split}
 \sum_{t=r}^s \big( g(x_t) b_t -   \lambda |g(x_t)-g(x_{t+1})| \big)  =&-O\left(\sqrt{(1+\lambda) \leh \log \frac{1}{Z}}   + \sqrt{(1+\lambda)} Z \tau\right) \\
\overset{Z=O(1/T)}{=}  &-O\left(\sqrt{(1+\lambda) \leh \log T} \right).
\end{split}
\end{equation}
for any interval $[r,s]$. Notice that the upper bound in (\ref{eqn:al2:prop1}) is independent of the interval length $\tau$, so it holds even when $\tau$ is \emph{larger} than $\leh$. On the other hand, for any interval $[r,s]$ whose length is no larger than $\leh$, i.e., $\tau/\leh\leq 1$, we also have the following regret bound of Algorithm~\ref{alg:2} w.r.t.~the baseline strategy which always outputs $1$:
\begin{equation}\label{eqn:al2:prop2}
\begin{split}
\sum_{t=r}^s \big( g(x_t) b_t -  \lambda |g(x_t)-g(x_{t+1})| \big)
\overset{Z=O(1/T)}{=}  \sum_{t=r}^s b_t  -O\left(\sqrt{(1+\lambda) \leh \log T} \right).
\end{split}
\end{equation}
As elaborated later, when using Algorithm~\ref{alg:2} to combine multiple algorithms, there are two issues that need to be addressed.
\begin{compactenum}
\item We do not destroy the theoretical guarantee of early algorithms, which is ensured by the property in (\ref{eqn:al2:prop1}).
\item We can inherit the theoretical guarantee of the current algorithm, which is archived by the property in (\ref{eqn:al2:prop2}).
\end{compactenum}

\subsection{Smoothed OGD}
We introduce the following common assumptions for OCO \citep{Online:suvery}.
\begin{ass}\label{ass:0} All the functions $f_t$'s are convex over the domain $\W$.
\end{ass}
\begin{ass}\label{ass:1} The gradients of all functions are bounded by $G$, i.e.,
\begin{equation}\label{eqn:grad}
\max_{\w \in \W}\|\nabla f_t(\w)\| \leq G, \ \forall t \in [T].
\end{equation}
\end{ass}
\begin{ass}\label{ass:2} The diameter of the domain $\W$ is bounded by $D$, i.e.,
\begin{equation}\label{eqn:domain}
\max_{\w, \w' \in \W} \|\w -\w'\| \leq D.
\end{equation}
\end{ass}
Without loss of generality, we assume
\begin{equation} \label{eqn:function:range}
f_t(\w) \in [0, GD], \ \forall \w \in \W, \ t \in [T],
\end{equation}
since we can always redefine $f_t(\w)$ as
\[
f_t(\w)-\min_{\w \in \W}f_t(\w)
\]
which belongs to $[0,GD]$ according to (\ref{eqn:grad}) and (\ref{eqn:domain}).

First, we demonstrate how to use DNP-cu to combine the predictions of two algorithms designed for OCO.  Let $\A^1$ and $\A^2$ be two online learners, and denote their predictions in the $t$-th round by $\w_t^1$ and $\w_t^2$, respectively. Let $\A$ be a meta-algorithm which outputs a convex combination of $\w_t^1$ and $\w_t^2$, i.e.,
\begin{equation} \label{eqn:weight:combin}
\w_t= (1-w_t) \w_t^1 +  w_t  \w_t^2
\end{equation}
where the weight $w_t\in[0,1]$. We have the following lemma regarding the meta-regret of $\A$ w.r.t.~$\A^1$ and $\A^2$.
\begin{lemma} \label{lem:combine:two} Assume the outputs of $\A^1$ and $\A^2$ move slowly such that
\begin{equation} \label{eqn:switch:expert}
\begin{split}
\|\w_t^1  - \w_{t+1}^1 \| \leq  &\frac{M D}{\lambda}, \\
 \|\w_t^2  - \w_{t+1}^2 \| \leq &\frac{MD}{\lambda}, \ \forall t \in [T]
 \end{split}
\end{equation}
where $M\geq0$ is some constant. Under Assumptions~\ref{ass:0}, \ref{ass:1} and \ref{ass:2}, the meta-regret of $\A$ w.r.t.~$\A^1$ over any interval $[r,s]$ satisfies
\begin{equation} \label{eqn:com:reg:1}
\begin{split}
& \sum_{t=r}^s \big( f_t(\w_t)+\lambda G\|\w_t-\w_{t+1}\|\big)  - \sum_{t=r}^s \big( f_t(\w_t^1)+\lambda G \|\w_t^1-\w_{t+1}^1\|\big) \\
\leq &  -(1+M)GD\sum_{t=r}^s  \big( w_t  \ell_t  - \lambda |w_t  -w_{t+1}| \big)
\end{split}
\end{equation}
and the meta-regret of $\A$ w.r.t.~$\A^2$ satisfies
\begin{equation} \label{eqn:com:reg:2}
\begin{split}
& \sum_{t=r}^s \big( f_t(\w_t)+\lambda G \|\w_t-\w_{t+1}\|\big)  - \sum_{t=r}^s \big( f_t(\w_t^2)+\lambda  G\|\w_t^2-\w_{t+1}^2\|\big) \\
\leq & -(1+M)GD  \sum_{t=r}^s \big( w_t \ell_t  - \lambda   |w_t  -w_{t+1}|  -  \ell_t \big)
\end{split}
\end{equation}
where
\begin{eqnarray} 
\ell_t^1= f_t(\w_t^1) +  \lambda G \|\w_t^1  - \w_{t+1}^1 \| &\overset{(\ref{eqn:function:range}),(\ref{eqn:switch:expert})}{\in}  &[0, (1+M)GD] ,\label{eqn:ell:1:range} \\
\ell_t^2 =f_t(\w_t^2) +  \lambda G\|\w_t^2  - \w_{t+1}^2 \|& \overset{(\ref{eqn:function:range}),(\ref{eqn:switch:expert})}{\in} & [0, (1+M)GD] , \label{eqn:ell:2:range} \\
\ell_t =\frac{\ell_t^1-\ell_t^2 }{(1+M)GD} &\overset{(\ref{eqn:ell:1:range}),(\ref{eqn:ell:2:range})}{\in} &[-1, 1]. \label{eqn:ell:definition}
\end{eqnarray}
\end{lemma}
\textbf{Remark: } Comparing Lemma~\ref{lem:combine:two} with Corollary~\ref{cor:1}, we immediately see that the weight $w_t$ in (\ref{eqn:weight:combin}) can be determined by invoking DNP-cu, i.e., Algorithm~\ref{alg:2} to process the scaled bit sequence
\[
\frac{\ell_1}{\max(\sqrt{\lambda},1)},\ldots, \frac{\ell_T}{\max(\sqrt{\lambda},1)}.
\]
Then, we can utilize lower bounds in Corollary~\ref{cor:1} to establish upper bounds for the regret in Lemma~\ref{lem:combine:two}. We name the strategy of aggregating two algorithms by DNP-cu as Combiner, and summarize its procedure in Algorithm~\ref{alg:3}.
\begin{algorithm}[t]
\caption{Combiner}
\begin{algorithmic}[1]
\REQUIRE Three parameters: $M$, $\leh$ and $Z$
\REQUIRE Two algorithms: $\A^1$ and $\A^2$
\STATE Let $\A$ be an instance of DNP-cu, i.e., Algorithm~\ref{alg:2}, with parameter $\leh$ and $Z$
\STATE Receive $\w_1^1$ and $\w_1^2$ from $\A^1$ and $\A^2$ respectively
\STATE Receive the prediction $w_{1}$ from $\A$
\FOR{$t=1,\ldots,T$}
\STATE Predict $\w_t$ according to (\ref{eqn:weight:combin})
\STATE Send the loss function $f_t(\cdot)$ to $\A^1$ and $\A^2$
\STATE Receive $\w_{t+1}^1$ and $\w_{t+1}^2$ from $\A^1$ and $\A^2$ respectively
\STATE Send the bit $\frac{\ell_{t}}{\max(\sqrt{\lambda},1)}$ to $\A$, where $\ell_t$ is defined in (\ref{eqn:ell:definition})
\STATE Receive the prediction $w_{t+1}$ from $\A$
\ENDFOR
\end{algorithmic}
\label{alg:3}
\end{algorithm}

In the following, we will use online gradient descent (OGD) with constant step size \citep{zinkevich-2003-online} as our expert-algorithm.  OGD performs gradient descent to update the current solution $\w_t$:
\[
\w_{t+1} = \Pi_{\W}\big[\w_t - \eta \nabla f_t(\w_t)\big]
\]
where $\eta>0$ is the step size, and $\Pi_{\W}[\cdot]$ denotes the projection onto $\W$. Notice that it is important to choose a \emph{constant} step size, which makes it easy to analyze the regret, as well as the dynamic regret, over any interval $[r,s]$.

First, we create $K$ instances of OGD, denoted by $\A^1,\ldots,\A^K$, where the value of $K$ will be determined later.  The step size of $\A^i$ is set to be
\begin{equation}\label{eqn:step:etai}
\eta^{(i)} = \frac{D}{G} \sqrt{\frac{1}{(1+ 2\lambda )\leh^{(i)}}}
\end{equation}
where
\begin{equation}\label{eqn:tau:value}
\leh^{(i)}= T 2^{1-i}.
\end{equation}
Then, we use Combiner, i.e., Algorithm~\ref{alg:3} to aggregate them sequentially. We will create a sequence of algorithms $\B^1,\ldots,\B^K$, where $\B^i$ is obtained by combining $\B^{i-1}$ with $\A^i$, and $\B^1=\A^1$. The parameter $\leh$ in Algorithm~\ref{alg:3} is set to be $\leh^{(i)}$ when forming $\B^i$. In each iteration $t$, we invoke $\B^1,\ldots,\B^K$ sequentially for one step, and return the output of $\B^K$ as the prediction $\w_t$. The completed procedure is named as smoothed OGD, and summarized in Algorithm~\ref{alg:4}.
\begin{algorithm}[t]
\caption{Smoothed OGD}
\begin{algorithmic}[1]
\REQUIRE Three parameters: $K$, $M$ and $Z$
\FOR{$i=1,\ldots,K$}
\STATE Set $\leh^{(i)} = T 2^{1-i}$
\STATE Let $\A^i$ be an instance of OGD with step size $\eta^{(i)}$ defined in (\ref{eqn:step:etai})
\IF{$i=1$}
\STATE Set $\B^1=\A^1$
\ELSE
\STATE Let $\B^i$ be an instance of Combiner, i.e., Algorithm~\ref{alg:3} which combines $B^{i-1}$ and $\A^i$ with parameters $M$, $\leh^{(i)}$ and $Z$
\ENDIF
\ENDFOR
\FOR{$t=1,\ldots,T$}
\STATE Run $\B^1,\ldots,\B^K$ sequentially for one step
\STATE Predict the output of $\B^K$, denoted by $\w_t$
\ENDFOR
\end{algorithmic}
\label{alg:4}
\end{algorithm}

\textbf{Remark: } Our method has a similar structure with those of \citet{pmlr-v119-cutkosky20a} and \citet{arXiv:2102.01623}, in the sense that we all combine multiple experts sequentially. In contrast, other approaches for adaptive regret use a two-level framework, where a meta-algorithm aggregates multiple experts (which is allowed to sleep) simultaneously \citep{Adaptive:Hazan,Adaptive:ICML:15,Improved:Strongly:Adaptive,jun2017,Adaptive:Regret:Smooth:ICML,Dual:Adaptivity}. On the other hand, all the previous works, including \citet{pmlr-v119-cutkosky20a} and \citet{arXiv:2102.01623}, need to construct a set of sub-intervals, and maintain a sub-routine for each one.  In this way, they can use a small number of sub-intervals to cover any possible interval, and attain a small regret on that interval. By comparison, our method does not rely on any special construction of sub-intervals, making it more elegant.

\subsection{Theoretical Guarantees}
Next, we provide the theoretical guarantee of smoothed OGD. We first characterize its regret with switching cost over any interval.
\begin{theorem} \label{thm:regret} Assume
\begin{equation} \label{eqn:T:lower}
T \geq \max( \sqrt{\lambda}\log_2 T, e)
\end{equation}
and set \begin{equation} \label{eqn:def:K}
K=\left\lfloor \log_2 \frac{T}{32 \max( \lambda, 1) \log 1/Z} \right\rfloor +1,
\end{equation}
$M=2$ and $Z=1/T$ in Algorithm~\ref{alg:4}. Under Assumptions~\ref{ass:0}, \ref{ass:1} and \ref{ass:2},  we have
\[
 \begin{split}
&\sum_{t=r}^s \big( f_t(\w_t)+  \lambda G  \|\w_t-\w_{t+1}\| - f_t(\w)\big)\\
\leq &2 GD \sqrt{(1+ \lambda)\tau}  +113 GD \max(\sqrt{\lambda},1) \sqrt{\tau \log T  } \\
=& O\left(\sqrt{(1+\lambda) \tau \log T }\right)
\end{split}
\]
for any interval $[r,s]$ with length $\tau$, and any $\w \in \W$.
\end{theorem}
\textbf{Remark: } Our $O(\sqrt{(1+\lambda) \tau \log T })$ bound is on the same order as that of \citet{arXiv:2102.01623}. When $\lambda=0$, we get the $O(\sqrt{\tau \log T})$ adaptive regret for general convex functions \citep{Improved:Strongly:Adaptive}.

Our proposed method is also equipped with nearly optimal dynamic regret with switching cost over any interval, as stated below.
\begin{theorem} \label{thm:dynamic:regret}  Under the condition of Theorem~\ref{thm:regret}, we have
\[
 \begin{split}
&\sum_{t=r}^s \big( f_t(\w_t)+  \lambda G  \|\w_t-\w_{t+1}\| - f_t(\u_t)\big)\\
\leq & 2 GD \sqrt{(1+ \lambda )\tau (1+2 P_{r,s}/D)} +120 GD \max(\sqrt{\lambda},1) \sqrt{ \tau (1+2P_{r,s}/D) \log T}\\
=& O\left(\sqrt{(1+\lambda) \tau (1+P_{r,s})\log T }\right)
\end{split}
\]
where $\tau$ is the interval length, and $P_{r,s}$, defined in (\ref{eqn:path:length:interval}), is the path-length of an arbitrary comparator sequence $\u_r,\ldots,\u_s\in \W$.
\end{theorem}
\textbf{Remark: } The above theorem shows that our method can minimize the dynamic regret with switching cost over any interval. According to the $\Omega(\sqrt{T(1+P_{1,T})})$ lower bound of dynamic regret \citep[Theorem 2]{Adaptive:Dynamic:Regret:NIPS}, we know that our upper bound is optimal, up to a logarithmic factor. Theorem~\ref{thm:dynamic:regret} is very general and can be simplified in different ways.
\begin{compactenum}
\item If we choose a fixed comparator such that $P_{r,s}=0$, Theorem~\ref{thm:dynamic:regret} reduces to Theorem~\ref{thm:regret} and matches that of \citet{arXiv:2102.01623}.
\item If we ignore the switching cost and set $\lambda=0$, we obtain an $O(\sqrt{\tau (1+P_{r,s})\log T })$ bound for dynamic regret over any interval, which recovers the results of \citet[Theorem 4]{Adaptive:Dynamic:AISTATS} and \citet[Theorem 7]{pmlr-v119-cutkosky20a}.
\item When both $P_{r,s}=0$ and $\lambda=0$, we obtain the $O(\sqrt{\tau \log T})$ adaptive regret of \citet{Improved:Strongly:Adaptive}.
\end{compactenum}

%
%
\section{Analysis}
In this section, we present the proof of all theorems.

\subsection{Proof of Theorem~\ref{thm:1}}
First, from the updating rule in Algorithm~\ref{alg:2}, we can prove that the derivation satisfies
\begin{equation}\label{eqn:xt:range}
-\mu \leq x_t  \leq U(\leh) + \mu, \ \forall t \geq 1.
\end{equation}
To see this, we first consider the upper bound in (\ref{eqn:xt:range}). Let $k$ be any iteration such that $x_k  \leq U(\leh)$ and $x_{k+1}  > U(\leh)$. Then, we must have $x_{k+1}=\rho x_k + b_k$, because otherwise $x_{k+1} =\rho x_k < U(\leh)$. As a result,
\[
x_{k+1} = \rho x_k + b_k \leq U(\leh) + \mu.
\]
Now, we consider the next derivation $x_{k+2}$. Because $x_{k+1}>U(\leh)$, according to the conservative updating rule, we have
\[
x_{k+2} = \left\{
            \begin{array}{ll}
              \rho x_{k+1} + b_{k+1}, & b_{k+1} < 0; \\
              \rho x_{k+1}, & \hbox{otherwise.}
            \end{array}
          \right.
\]
which is always smaller than $x_{k+1}$. Repeating the above argument, we conclude that the subsequent derivations  $x_{k+2},x_{k+3},\ldots$ keep decreasing until they become no bigger than $U(\leh)$. As a result, it is impossible for $x_t$ to exceed $U(\leh) + \mu$.

The lower bound in (\ref{eqn:xt:range}) can be proved in a similar way. Let $k$ be any iteration such that $x_k \geq0 $ and $x_{k+1}  <0 $. Then, we must have $x_{k+1}=\rho x_k + b_k$, because otherwise $x_{k+1} =\rho x_k \geq 0 $. As a result,
\[
x_{k+1} = \rho x_k + b_k \geq  b_k  \geq -\mu.
\]
Now, we consider the next derivation $x_{k+2}$. Because $x_{k+1}<0$, according to the conservative updating rule, we have
\[
x_{k+2} = \left\{
            \begin{array}{ll}
              \rho x_{k+1} + b_{k+1}, & b_{k+1} > 0; \\
              \rho x_{k+1}, & \hbox{otherwise.}
            \end{array}
          \right.
\]
which is always bigger than $x_{k+1}$. Repeating the above argument, we conclude that the subsequent derivations  $x_{k+2},x_{k+3},\ldots$ keep increasing until they become nonnegative. As a result, it is impossible for $x_t$ to be smaller than $-\mu$.

Next, we make use of Algorithm~\ref{alg:1} to analyze the reward of Algorithm~\ref{alg:2}. Following \citet{DNP:2010:Arxiv}, we construct the following bit sequence
\[
\tb_t = \left\{
            \begin{array}{ll}
              b_t, & \hbox{if Line 6 of Algorithm~\ref{alg:2} is executed at round $t$}; \\
              0, & \hbox{otherwise.}
            \end{array}
          \right.
\]
It is easy to verify that the prediction $g(x_t)$, as well as the derivation $x_t$, of Algorithm~\ref{alg:2} over the bit sequence $b_1,\ldots,b_T$ is exact the same as that of Algorithm~\ref{alg:1} over the new sequence $\tb_1,\ldots,\tb_T$. Since Algorithm~\ref{alg:1} is more simple, we will first establish the theoretical guarantee of Algorithm~\ref{alg:1} over the new sequence, and then convert it to the reward of Algorithm~\ref{alg:2} over the original sequence. We have the following theorem for Algorithm~\ref{alg:1} \citep{pmlr-v98-daniely19a}.

\begin{theorem} \label{thm:2} Suppose $Z \leq \frac{1}{e}$ and $\leh \geq \max\{8e, 16\log \frac{1}{Z}\}$. For any bit sequence $b_1,\ldots,b_T$ such that $|b_t| \leq \mu \leq 1$,  the cumulative reward of Algorithm~\ref{alg:1} over any interval $[r,s]$ with length $\tau$ satisfies
\begin{equation} \label{eqn:alg1:lower}
\begin{split}
&\sum_{t=r}^s \left( g(x_t) b_t -   \frac{1}{\bb} |g(x_t)-g(x_{t+1})| \right) \\
\geq &\max\left(0, \sum_{t=r}^s b_t + x_r  - \frac{\tau}{\leh} \left( U(\leh) +2 \mu\right) - U(\leh) \right)  -  \max(x_r,0)- Z \tau
\end{split}
\end{equation}
where $U(\leh)$ is defined in (\ref{eqn:u:tau}). And the change of successive predictions satisfies
\begin{equation} \label{eqn:change:prediction:F}
|g(x_t)-g(x_{t+1})| \leq  \bb  \sqrt{\frac{1}{\leh} \log \frac{1}{Z}}+ \frac{Z \bb}{4}.
\end{equation}
\end{theorem}
Theorem~\ref{thm:2} can be extracted from the proofs of Lemmas 21 and 23 of \citet{pmlr-v98-daniely19a}. For the sake of completeness, we provide its analysis in Appendix~\ref{sec:proof:thm2}. We can see that the lower bound in (\ref{eqn:alg1:lower}) depends on $x_r$, which explains the necessity of controlling its value.

Notice that $\mu$ is also the upper bound of the absolute value of the new sequence $\tb_1,\ldots,\tb_T$. According to Theorem~\ref{thm:2}, we directly obtain (\ref{eqn:change:prediction:F:2}) from (\ref{eqn:change:prediction:F}). From (\ref{eqn:alg1:lower}), we have
\begin{equation} \label{eqn:new:reward}
\begin{split}
&\sum_{t=r}^s \left( g(x_t) \tb_t -   \frac{1}{\bb} |g(x_t)-g(x_{t+1})| \right)\\
\geq &\max\left(0, \sum_{t=r}^s \tb_t + x_r  - \frac{\tau}{\leh} \left( U(\leh) +2 \mu\right) - U(\leh) \right)  -  \max(x_r,0)- Z \tau.
\end{split}
\end{equation}
On the other hand, the reward in terms of the original sequence is
\begin{equation}  \label{eqn:diff:reward}
\begin{split}
&\sum_{t=r}^s \left(g(x_t) b_t -   \frac{1}{\bb} |g(x_t)-g(x_{t+1})| \right)\\
=& \sum_{t=r}^s g(x_t) (b_t -  \tb_t) + \sum_{t=r}^s \left(g(x_t) \tb_t -   \frac{1}{\bb} |g(x_t)-g(x_{t+1})| \right).
\end{split}
\end{equation}
So, we need to bound $\sum_{t=r}^s g(x_t) (b_t -  \tb_t) $. Let $k$ be any iteration such that $b_k \neq \tb_k$, i.e., Line 8 of Algorithm~\ref{alg:2} is executed at round $k$, which also implies $\tb_k=0$. From the updating rule, we must have
\[
x_k <0 \cap  b_k \leq 0 \textrm{ or }x_k >U(\leh) \cap b_k \geq 0.
\]
If $x_k <0 \cap  b_k \leq 0$, we have
\[
g(x_k) (b_k -  \tb_k) = g(x_k) b_k = 0 \geq b_k =b_k-\tb_k
\]
since $g(x_k)=0$ and $\tb_k=0$. Otherwise if $x_k >U(\leh) \cap b_k \geq 0$, we have
\[
g(x_k) (b_k -  \tb_k) = b_k = b_k-\tb_k \geq 0
\]
since  $g(x_k)=1$ and $\tb_k=0$. So, we always have
\begin{equation}  \label{eqn:diff:payoff}
g(x_k) (b_k -  \tb_k) \geq \max\left(0, b_k-\tb_k\right), \textrm{ if } b_k \neq \tb_k.
\end{equation}
As a result,
\begin{equation}  \label{eqn:diff:bound}
\begin{split}
&\sum_{t=r}^s g(x_t) (b_t -  \tb_t) =  \sum_{t \in [r,s] \cap b_t \neq  \tb_t} g(x_t) (b_t -  \tb_t) \\
\overset{(\ref{eqn:diff:payoff})}{\geq} & \max\left(0, \sum_{t \in [r,s] \cap b_t \neq  \tb_t} \left(b_t-\tb_t \right) \right)=\max\left(0,\sum_{t=r}^s \left(b_t-\tb_t \right) \right).
\end{split}
\end{equation}

Combining (\ref{eqn:new:reward}), (\ref{eqn:diff:reward}) and (\ref{eqn:diff:bound}), we have
\[
\begin{split}
&\sum_{t=r}^s \left(g(x_t) b_t -   \frac{1}{\bb} |g(x_t)-g(x_{t+1})| \right)\\
\geq & \max\left(0, \sum_{t=r}^s \tb_t + x_r  - \frac{\tau}{\leh} \left( U(\leh) +2 \mu\right) - U(\leh) \right)  -  \max(x_r,0)- Z \tau
\\
&+ \max\left(0,\sum_{t=r}^s \left(b_t-\tb_t \right) \right) \\
\geq & \max\left(0, \sum_{t=r}^s b_t + x_r  - \frac{\tau}{\leh} \left( U(\leh) +2 \mu\right) - U(\leh) \right)  -  \max(x_r,0)- Z \tau.
\end{split}
\]
Then, we obtain (\ref{eqn:alg2:lower}) by using (\ref{eqn:xt:range}) to bound $x_r$, and obtain (\ref{eqn:alg2:lower:second}) based on $x_1=0$.
\subsection{Proof of Corollary~\ref{cor:1}}
Notice that the magnitude of the scaled bit sequence is upper bounded by $\mu=1/\max(\sqrt{\lambda},1)$. From Theorem~\ref{thm:1}, we have
\begin{equation} \label{eqn:cor:1:1}
\begin{split}
&\sum_{t=r}^s \left( g(x_t) \frac{b_t}{\max(\sqrt{\lambda},1)} -   \max(\sqrt{\lambda},1)  |g(x_t)-g(x_{t+1})| \right)\\
\overset{(\ref{eqn:alg2:lower})}{\geq}  &\max\left(0, \sum_{t=r}^s \frac{b_t}{\max(\sqrt{\lambda},1)}    - \frac{\tau}{\leh} \left( U(\leh) +\frac{2}{\max(\sqrt{\lambda},1)} \right) - U(\leh) -\frac{1}{\max(\sqrt{\lambda},1)} \right) \\
& -  U(\leh) -\frac{1}{\max(\sqrt{\lambda},1)}  - Z\tau.
\end{split}
\end{equation}
Then, we can lower bound the cumulative reward as follows
\[
\begin{split}
&\sum_{t=r}^s \big( g(x_t) b_t - \lambda |g(x_t)-g(x_{t-1})| \big) \\
\geq & \max(\sqrt{\lambda},1) \sum_{t=r}^s \left( g(x_t) \frac{b_t}{\max(\sqrt{\lambda},1)} -   \max(\sqrt{\lambda},1)  |g(x_t)-g(x_{t+1})| \right)\\
\overset{(\ref{eqn:cor:1:1})}{\geq} &\max\left(0, \sum_{t=r}^s b_t   - \max(\sqrt{\lambda},1) U(\leh) \left(\frac{\tau}{\leh} +1 \right)- \frac{2\tau}{\leh}  -1 \right)
 - \max(\sqrt{\lambda},1) U(\leh) \\
 & -1  -\max(\sqrt{\lambda},1) Z \tau
\end{split}
\]
which proves (\ref{eqn:cor:lower}). The upper bound in  (\ref{eqn:change:prediction:F:3}) is a direct consequence of (\ref{eqn:change:prediction:F:2}).

\subsection{Proof of Theorem~\ref{thm:regret}}
First, we show that under our setting of parameters, all the preconditions in Corollary~\ref{cor:1} and Lemma~\ref{lem:combine:two} are satisfied so that they can be exploited to analyze $\B^i$, which invokes Algorithm~\ref{alg:2} to combine $\B^{i-1}$ and $\A^i$. From (\ref{eqn:T:lower}), we know that $Z=1/T \leq 1/e$. From our definition of $K$, we have
\begin{equation} \label{eqn:lower:tau}
\leh^{(i)} \geq T 2^{1-K} \overset{(\ref{eqn:def:K})}{\geq}32 \max( \lambda, 1) \log \frac{1}{Z} \geq 32 \log \frac{1}{Z}  \geq 32 \geq 8e  , \ \forall i \in [K].
\end{equation}
Thus, the conditions about $Z$ and $\leh$ in Corollary~\ref{cor:1} are satisfied. Furthermore, our choice of $M$ ensures that (\ref{eqn:switch:expert}) in Lemma~\ref{lem:combine:two} is true. To this end, we prove the following lemma.
\begin{lemma} \label{lem:condition:M}  For all $\A^i$'s and $\B^i$'s created in Algorithm~\ref{alg:4}, their outputs satisfy the condition in (\ref{eqn:switch:expert}) with $M=2$.
\end{lemma}
Based on above discussions, we conclude that Corollary~\ref{cor:1} and Lemma~\ref{lem:combine:two} can be used in our analysis.

Next, we introduce the following theorem about the regret of OGD with switching cost over any interval $[r,s]$, which will be used to analyze the performance of $\A^i$'s.
\begin{theorem} \label{thm:regret:ogd} Let $\x_t$ be the outputs of OGD with step size $\eta$. Under Assumptions~\ref{ass:0}, \ref{ass:1} and \ref{ass:2}, we have
\[
\sum_{t=r}^s \big( f_t(\w_t)+  \lambda G  \|\w_t-\w_{t+1}\| - f_t(\w)\big) \leq\frac{D^2}{2 \eta} +   \frac{(1+ 2\lambda )\eta(s-r+1)  G^2}{2}
\]
for any $\w \in \W$.
\end{theorem}

\paragraph{Long Intervals} We proceed to analyze the performance of Algorithm~\ref{alg:4} over an interval $[r,s]$, and start with the case that the interval length
\[
\tau=s-r+1 \geq 32 \max( \lambda, 1) \log \frac{1}{Z}.
\]
From our construction of $\leh^{(i)}$ in (\ref{eqn:tau:value}), there must exist a
\begin{equation} \label{eqn:value:k}
k = \left\lfloor \log_2 \frac{T}{\tau} \right\rfloor +1 \leq K
\end{equation}
such that
\begin{equation} \label{eqn:tauk:property}
 \frac{\leh^{(k)}}{2} \leq \tau \leq \leh^{(k)}.
\end{equation}
Then, we divide the proof into two steps:
\begin{compactenum}[(i)]
  \item We show that the algorithm $\A^k$ attains an optimal regret with switching cost over the interval $[r,s]$;
  \item We demonstrate that the regret of $\B^K$ w.r.t.~$\A^k$ is under control.
\end{compactenum}

Let $\w_t^k$ be the output of $\A^k$ in the $t$-th iteration. From Theorem~\ref{thm:regret:ogd}, we have
\begin{equation} \label{eqn:bound:A^k}
\begin{split}
&\sum_{t=r}^s \big( f_t(\w_t^k)+  \lambda G  \|\w_t^k-\w_{t+1}^k\| - f_t(\w)\big) \\
\leq & \frac{D^2}{2 \eta^{(k)}} +   \frac{(1+ 2\lambda )\eta^{(k)}  \tau  G^2}{2} \overset{(\ref{eqn:step:etai})}{=}\frac{GD}{2} \sqrt{(1+ 2\lambda )\leh^{(k)}} + \frac{   GD  }{2}\tau \sqrt{\frac{1+ 2\lambda }{\leh^{(k)}}} \\
\overset{(\ref{eqn:tauk:property})}{\leq } & \frac{(\sqrt{2}+1)GD}{2} \sqrt{(1+ 2\lambda )\tau} \leq 2 GD \sqrt{(1+ \lambda )\tau}.
\end{split}
\end{equation}

Let $\bv_t^i$ be the output of $\B^i$ in the $t$-th iteration. We establish the following lemma to bound the regret of $\B^K$ w.r.t.~$\A^k$.
\begin{lemma} \label{lem:BK:Bk} For any interval $[r,s]$ with length $\tau \leq c \leh^{(k)}$, we have
\begin{equation}\label{eqn:BK:Bk:1}
\begin{split}
&\sum_{t=r}^s \big( f_t(\bv_t^K)+  \lambda G  \|\bv_t^K-\bv_{t+1}^K\|\big) - \sum_{t=r}^s \big( f_t(\w_t^k)+  \lambda G  \|\w_t^k-\w_{t+1}^k\|\big) \\
\leq & GD \max(\sqrt{\lambda},1) \left( (12c+53) \sqrt{\leh^{(k)} \log T}  +9+6c  + 6 (K-k)\right).
\end{split}
\end{equation}
\end{lemma}
Based on Lemma~\ref{lem:BK:Bk}, we have
\begin{equation} \label{eqn:BK:Bk:2}
\begin{split}
&\sum_{t=r}^s \big( f_t(\bv_t^K)+  \lambda G  \|\bv_t^K-\bv_{t+1}^K\|\big) - \sum_{t=r}^s \big( f_t(\w_t^k)+  \lambda G  \|\w_t^k-\w_{t+1}^k\|\big) \\
\overset{(\ref{eqn:tauk:property}),(\ref{eqn:BK:Bk:1})}{\leq } &   GD \max(\sqrt{\lambda},1) \left( 65 \sqrt{2 \tau \log T} +15  \right) + 6  GD \max(\sqrt{\lambda},1) (K-k) \\
\leq & 107 GD \max(\sqrt{\lambda},1)  \sqrt{\tau \log T} +  6  GD \max(\sqrt{\lambda},1) \log_2 \tau\\
\overset{\log_2 \tau \leq \sqrt{\tau \log \tau}}{\leq} & 113 GD \max(\sqrt{\lambda},1)  \sqrt{\tau \log T}
\end{split}
\end{equation}
where in the penultimate step we make use of the following fact
\[
\begin{split}
K-k \overset{(\ref{eqn:def:K}), (\ref{eqn:value:k})}{= }&\left\lfloor \log_2 \frac{T}{32 \max( \lambda, 1) \log 1/Z} \right\rfloor - \left\lfloor \log_2 \frac{T}{\tau} \right\rfloor \\
\leq &\log_2 \frac{\tau}{32 \max( \lambda, 1) \log 1/Z} +1 \leq \log_2 \tau.
\end{split}
\]
Combining (\ref{eqn:bound:A^k}) and (\ref{eqn:BK:Bk:2}), we have
\begin{equation} \label{eqn:bound:long:interval}
\begin{split}
&\sum_{t=r}^s \big( f_t(\w_t^K)+  \lambda G  \|\w_t^K-\w_{t+1}^K\| - f_t(\w)\big)\\
\leq & 2 GD \sqrt{(1+ \lambda )\tau}  +113 GD \max(\sqrt{\lambda},1) \sqrt{\tau \log T  } .
\end{split}
\end{equation}

\paragraph{Short Intervals} We study short intervals $[r,s]$ such that
\[
\tau=s-r+1 \leq 32 \max( \lambda, 1) \log \frac{1}{Z}.
\]
From Lemma~\ref{lem:condition:M}, we know that the output of $B^K$ moves slowly such that
\begin{equation} \label{eqn:movement:BK}
\|\w_t^K-\w_{t+1}^K\|  \leq \frac{2 D}{\lambda}.
\end{equation}
As a result, the regret of $B^K$ over $[r,s]$ can be bounded by
\begin{equation} \label{eqn:bound:short:interval}
\begin{split}
&\sum_{t=r}^s \big( f_t(\w_t^K)+  \lambda G  \|\w_t^K-\w_{t+1}^K\| - f_t(\w)\big) \leq \big( f_t(\w_t^K)+  \lambda G  \|\w_t^K-\w_{t+1}^K\|\big) \\
\overset{(\ref{eqn:function:range}), (\ref{eqn:movement:BK})}{\leq} & 3 \tau GD \leq  3 GD \sqrt{ \tau \cdot 32 \max( \lambda, 1) \log T }  \leq  17 GD \max(\sqrt{\lambda},1) \sqrt{\tau \log T}.
\end{split}
\end{equation}

We complete the proof by combing (\ref{eqn:bound:long:interval}) and (\ref{eqn:bound:short:interval}).

\subsection{Proof of Theorem~\ref{thm:dynamic:regret}}
Since we focus on dynamic regret, so we need the following theorem regarding the dynamic regret of OGD with switching cost over any interval $[r,s]$.
\begin{theorem} \label{thm:dynamic:regret:ogd} Under Assumptions~\ref{ass:0}, \ref{ass:1} and \ref{ass:2}, we have
\[
\sum_{t=r}^s \big( f_t(\w_t)+  \lambda G  \|\w_t-\w_{t+1}\| - f_t(\u_t)\big) \leq \frac{D^2}{2 \eta}+ \frac{D}{\eta} \sum_{t=r}^s \|\u_t-\u_{t+1}\|_2 + \frac{(1+ 2\lambda )\eta(s-r+1)  G^2}{2}
\]
for any comparator sequence $\u_r,\ldots,\u_s \in \W$.
\end{theorem}

The proof is similar to that of Theorem~\ref{thm:regret}, and  we consider two scenarios: long intervals and short intervals. Here, we multiply the interval length $\tau$ by $1/(1+2P_{r,s}/D)$ to reflect the fact that the comparator is changing.

\paragraph{Long Intervals} First, we study the case that
\[
 \frac{\tau}{1+2P_{r,s}/D} \geq 32 \max( \lambda, 1) \log \frac{1}{Z}.
\]
From our construction of $\leh^{(i)}$ in (\ref{eqn:tau:value}), there must exist a
\begin{equation} \label{eqn:value:k:dynamic}
k = \left\lfloor \log_2 \frac{T(1+2P_{r,s}/D)}{\tau} \right\rfloor +1 \leq K
\end{equation}
such that
\begin{equation} \label{eqn:tauk:property:dynamic}
 \frac{\leh^{(k)}}{2} \leq \frac{\tau}{1+2P_{r,s}/D}  \leq \leh^{(k)}.
\end{equation}
Next, we show that the dynamic regret of $\A^k$ with switching cost is almost optimal. From Theorem~\ref{thm:dynamic:regret:ogd}, we have
\begin{equation} \label{eqn:Dynamic:bound:A^k}
\begin{split}
&\sum_{t=r}^s \big( f_t(\w_t^k)+  \lambda G  \|\w_t^k-\w_{t+1}^k\| - f_t(\u_t)\big) \\
\leq & \frac{D^2}{2 \eta^{(k)}}+ \frac{D}{\eta^{(k)}} P_{r,s}  + \frac{(1+ 2\lambda )\eta^{(k)} \tau  G^2}{2}\\
\overset{(\ref{eqn:step:etai})}{=} & \frac{G(D+ 2 P_{r,s})}{2 } \sqrt{(1+ 2\lambda )\leh^{(k)}}   + \frac{GD \tau }{2}\sqrt{\frac{1+ 2\lambda}{\leh^{(k)}}}\\
\overset{(\ref{eqn:tauk:property:dynamic})}{\leq} &  \frac{(\sqrt{2}+1)GD}{2} \sqrt{(1+ 2\lambda )\tau (1+2 P_{r,s}/D)} \leq 2 GD \sqrt{(1+ \lambda )\tau (1+2 P_{r,s}/D)}.
\end{split}
\end{equation}

Then, we prove that the regret of $\B^K$ w.r.t.~$\A^k$ is roughly on the same order as (\ref{eqn:Dynamic:bound:A^k}).
From Lemma~\ref{lem:BK:Bk}, we have
\begin{equation} \label{eqn:Dynamic:bound:B^K}
\begin{split}
&\sum_{t=r}^s \big( f_t(\bv_t^K)+  \lambda G  \|\bv_t^K-\bv_{t+1}^K\|\big) - \sum_{t=r}^s \big( f_t(\w_t^k)+  \lambda G  \|\w_t^k-\w_{t+1}^k\|\big) \\
\overset{(\ref{eqn:tauk:property:dynamic}),(\ref{eqn:BK:Bk:1})}{\leq } &   GD \max(\sqrt{\lambda},1) \left( (65 + 24P_{r,s}/D  ) \sqrt{\leh^{(k)} \log T} +15+12P_{r,s}/D \right)
+ \\
&6  GD \max(\sqrt{\lambda},1) (K-k) \\
\overset{(\ref{eqn:tauk:property:dynamic})}{\leq } &   GD \max(\sqrt{\lambda},1) \left( 65 \sqrt{2 \tau (1+2P_{r,s}/D) \log T} +15+12P_{r,s}/D \right) \\
&+ 6  GD \max(\sqrt{\lambda},1) (K-k) \\
\leq & 114 GD \max(\sqrt{\lambda},1) \sqrt{ \tau (1+2P_{r,s}/D) \log T} + 6  GD \max(\sqrt{\lambda},1) \log_2 \tau\\
\overset{\log_2 \tau \leq \sqrt{\tau \log \tau}}{\leq} &  120 GD \max(\sqrt{\lambda},1) \sqrt{ \tau (1+2P_{r,s}/D) \log T}
\end{split}
\end{equation}
where in the penultimate step we use the following inequalities
\[
\begin{split}
15+12P_{r,s}/D & \overset{P_{r,s} \leq \tau D}{\leq } 15+12 \sqrt{\tau P_{r,s}/D} \overset{a+b\leq \sqrt{2a^2+2b^2}}{\leq }  15 \sqrt{2 \tau (1+2P_{r,s}/D)},\\
K-k &\overset{(\ref{eqn:def:K}), (\ref{eqn:value:k:dynamic})}{= }\left\lfloor \log_2 \frac{T}{32 \max( \lambda, 1) \log 1/Z} \right\rfloor - \left\lfloor \log_2 \frac{T(1+2P_{r,s}/D)}{\tau} \right\rfloor \leq \log_2 \tau.
\end{split}
\]
Combining (\ref{eqn:Dynamic:bound:A^k}) and (\ref{eqn:Dynamic:bound:B^K}), we can bound the dynamic regret of $\B^K$ with switching cost by
\begin{equation} \label{eqn:dynamic:bound:long:interval}
\begin{split}
&\sum_{t=r}^s \big( f_t(\w_t^K)+  \lambda G  \|\w_t^K-\w_{t+1}^K\| - f_t(\u_t)\big)\\
\leq & 2 GD \sqrt{(1+ \lambda )\tau (1+2 P_{r,s}/D)} +120 GD \max(\sqrt{\lambda},1) \sqrt{ \tau (1+2P_{r,s}/D) \log T}.
\end{split}
\end{equation}

\paragraph{Short Intervals} We consider short intervals $[r,s]$ such that
\[
 \frac{\tau}{1+2P_{r,s}/D} \geq 32 \max( \lambda, 1) \log \frac{1}{Z}.
\]
Following the analysis of Theorem~\ref{thm:regret}, the dynamic regret of $B^K$ over $[r,s]$ can be bounded by
\begin{equation} \label{eqn:dynamic:bound:short:interval}
\begin{split}
&\sum_{t=r}^s \big( f_t(\w_t^K)+  \lambda G  \|\w_t^K-\w_{t+1}^K\| - f_t(\u_t)\big)\\
\leq &3 \tau GD \leq  3 GD \sqrt{ \tau \cdot 32 \max( \lambda, 1) \log T  \cdot (1+2P_{r,s}/D)}  \\
\leq & 17 GD \max(\sqrt{\lambda},1) \sqrt{\tau (1+2P_{r,s}/D) \log T}.
\end{split}
\end{equation}

We complete the proof by combing (\ref{eqn:dynamic:bound:long:interval}) and (\ref{eqn:dynamic:bound:short:interval}).
\subsection{Proof of Theorem~\ref{thm:regret:ogd}}
From the standard analysis of OGD \citep{zinkevich-2003-online}, we have the following regret bound
\begin{equation} \label{eqn:reg:ogd}
\sum_{t=r}^s  \big( f_t(\w_t) - f_t(\w) \big) \leq \frac{D^2}{2 \eta} + \frac{\eta (s-r+1) G^2}{2}.
\end{equation}
To bound the switching cost, we have
\begin{equation} \label{eqn:switch:ogd}
\begin{split}
&\sum_{t=r}^s  \|\w_t-\w_{t+1}\| = \sum_{t=r}^s  \left\|\w_{t}-\Pi_{\W}\big[\w_{t} - \eta \nabla f_{t}(\w_{t})\big]\right\| \\
\leq &  \sum_{t=r}^s  \left\|- \eta \nabla f_{t}(\w_{t}) \right\| = \eta \sum_{t=r}^s  \left\|\nabla f_{t}(\w_{t}) \right\|  \overset{(\ref{eqn:grad})}{\leq} \eta (s-r+1) G.
\end{split}
\end{equation}
From (\ref{eqn:reg:ogd}) and (\ref{eqn:switch:ogd}), we have
\[
\sum_{t=r}^s \big( f_t(\w_t)+  \lambda G  \|\w_t-\w_{t+1}\| - f_t(\w)\big)\leq \frac{D^2}{2 \eta} + \frac{\eta (s-r+1) G^2}{2} +  \lambda \eta (s-r+1) G^2.
\]

\subsection{Proof of Theorem~\ref{thm:dynamic:regret:ogd}}
From the dynamic regret of OGD \citep{zinkevich-2003-online}, in particular Theorem 6 of \citet{Adaptive:Dynamic:Regret:NIPS}, we have
\[
\sum_{t=r}^s \big( f_t(\w_t)- f_t(\u_t)\big) \leq \frac{D^2}{2 \eta}+ \frac{D}{\eta} \sum_{t=r}^s \|\u_t-\u_{t+1}\|_2 + \frac{\eta (s-r+1)}{2 } G^2.
\]
We complete the proof by combining the above inequality with (\ref{eqn:switch:ogd}).

\subsection{Proof of Lemma~\ref{lem:combine:two}}
Similar to the analysis of Theorem 22 of \citet{pmlr-v98-daniely19a}, we decompose the weighted sum of hitting cost and switching cost as
\begin{equation} \label{eqn:decmpos:loss}
\begin{split}
&f_t(\w_t)+\lambda G\|\w_t-\w_{t+1}\| \\
=& f_t\left((1-w_t)\w_t^1 + w_t  \w_t^2\right)+\lambda G\left \| (1-w_t)\w_t^1 +  w_t\w_t^2- (1-w_{t+1}) \w_{t+1}^1 -  w_{t+1}\w_{t+1}^2 \right\| \\
\leq &  (1-w_t) f_t(\w_t^1) + w_t f_t(\w_t^2)+\lambda G\left\| (1-w_t)(\w_t^1  - \w_{t+1}^1) \right\| + \lambda G\left \| w_t(\w_t^2 -\w_{t+1}^2) \right \| \\
&+  \lambda G\left\| (1-w_t) \w_{t+1}^1 - (1-w_{t+1})\w_{t+1}^1 + w_t \w_{t+1}^2 -  w_{t+1}\w_{t+1}^2 \right\| \\
=&  (1-w_t)\left(f_t(\w_t^1) +  \lambda G \|\w_t^1  - \w_{t+1}^1 \|\right) + w_t\left( f_t(\w_t^2) + \lambda G \|\w_t^2 -\w_{t+1}^2 \|\right)\\
&+  \lambda G \left\| (w_t  -w_{t+1}) (\w_{t+1}^1  - \w_{t+1}^2 ) \right\| \\
\overset{(\ref{eqn:domain})}{\leq}  & (1-w_t)\left(f_t(\w_t^1) +  \lambda G \|\w_t^1  - \w_{t+1}^1 \|\right) + w_t\left( f_t(\w_t^2) + \lambda  G \|\w_t^2 -\w_{t+1}^2 \|\right) \\
&+  \lambda G D |w_t  -w_{t+1}|.
\end{split}
\end{equation}

Then, the regret of $\A$ w.r.t.~$\A^1$ over any interval $[r,s]$ can be upper bounded in the following way:
\[
\begin{split}
& \sum_{t=r}^s \big( f_t(\w_t)+\lambda G\|\w_t-\w_{t+1}\|\big) - \sum_{t=r}^s \big( f_t(\w_t^1)+\lambda G\|\w_t^1-\w_{t+1}^1\|\big) \\
\overset{(\ref{eqn:decmpos:loss})}{\leq} &  \sum_{t=r}^s  \Big( w_t \left[ \left( f_t(\w_t^2) + \lambda G \|\w_t^2 -\w_{t+1}^2 \|\right)-\left( f_t(\w_t^1) +  \lambda G \|\w_t^1  - \w_{t+1}^1 \| \right)\right]  \\
&+ \lambda G D |w_t  -w_{t+1}| \Big)\\
\overset{(\ref{eqn:ell:1:range}),(\ref{eqn:ell:2:range})}{=}&  \sum_{t=r}^s  \left( w_t (\ell_t^2 -\ell_t^1 )  + \lambda G D |w_t  -w_{t+1}| \right) \\
\overset{(\ref{eqn:ell:definition})}{=} & -(1+M)GD \sum_{t=r}^s  \left(w_t  \ell_t  -\frac{\lambda}{1+M}  |w_t  -w_{t+1}| \right)\\
\leq &  -(1+M)GD \sum_{t=r}^s \big( w_t  \ell_t  - \lambda |w_t  -w_{t+1}| \big)\\
\end{split}
\]
which proves (\ref{eqn:com:reg:1}). Similarly, the regret of $\A$ w.r.t.~$\A^2$ over any interval $[r,s]$ can be upper bounded by
\[
\begin{split}
& \sum_{t=r}^s \big( f_t(\w_t)+\lambda G \|\w_t-\w_{t+1}\|\big) - \sum_{t=r}^s \big( f_t(\w_t^2)+\lambda  G\|\w_t^2-\w_{t+1}^2\|\big) \\
\overset{(\ref{eqn:decmpos:loss})}{\leq} &\sum_{t=r}^s  (1-w_t) \left[ \left( f_t(\w_t^1) +  \lambda G\|\w_t^1  - \w_{t+1}^1 \| \right) -\left( f_t(\w_t^2) + \lambda G \|\w_t^2 -\w_{t+1}^2 \|\right)\right]\\
& + \sum_{t=r}^s \lambda GD |w_t  -w_{t+1}| \\
\overset{(\ref{eqn:ell:1:range}),(\ref{eqn:ell:2:range})}{=} & \sum_{t=r}^s  \left( (1-w_t) (\ell_t^1 - \ell_t^2 )  + \lambda GD |w_t  -w_{t+1}|\right) \\
\overset{(\ref{eqn:ell:definition})}{=} & - (1+M)GD  \sum_{t=r}^s \left( w_t \ell_t  -\frac{\lambda}{1+M}   |w_t  -w_{t+1}|  -  \ell_t \right)\\
\leq & -(1+M)GD  \sum_{t=r}^s \big( w_t \ell_t  - \lambda   |w_t  -w_{t+1}|  -  \ell_t \big)
\end{split}
\]
which proves (\ref{eqn:com:reg:2}).

\subsection{Proof of Lemma~\ref{lem:condition:M}}
We will prove that the outputs of $\A^i$'s and $\B^i$'s move slowly such that (\ref{eqn:switch:expert})  holds.
Let $\w_t^i$ be the output of $\A^i$ in the $t$-th iteration. From the updating rule of OGD, we have
\begin{equation} \label{eqn:movement:OGD}
 \|\w_t^i-\w_{t+1}^i\| \leq   \eta^{(i)} \left\|\nabla f_{t}(\w_t^i) \right\|  \overset{(\ref{eqn:grad})}{\leq}   \eta^{(i)}  G \overset{(\ref{eqn:step:etai})}{=}  D \sqrt{\frac{1}{(1+ 2\lambda )\leh^{(i)}}} \overset{(\ref{eqn:lower:tau})}{\leq} \frac{D}{\lambda}.
\end{equation}
So, $\w_t^i$'s satisfy the condition in (\ref{eqn:switch:expert}) when $M=2$.

Let $\bv_t^i$ be the output of $\B^i$ in the $t$-th iteration. We will prove by induction that
\begin{equation}\label{eqn:movement:BiS}
 \|\bv_t^i-\bv_{t+1}^i\| \leq \frac{D}{\lambda} + \frac{D}{\max(\sqrt{\lambda},1)}  \sum_{j=2}^{i} \left( \sqrt{\frac{1}{\leh^{(j)}} \log \frac{1}{Z}}+ \frac{Z}{4} \right), \ \forall i \in[K].
\end{equation}
The above equation, together with the following fact
\begin{equation} \label{eqn:upper:movement}
\begin{split}
& \frac{D}{\lambda} + \frac{D}{\max(\sqrt{\lambda},1)}  \sum_{j=2}^{K} \left( \sqrt{\frac{1}{\leh^{(j)}} \log \frac{1}{Z}}+ \frac{Z}{4} \right) \\
\overset{(\ref{eqn:tau:value})}{=}& \frac{D}{\lambda} + \frac{D}{\max(\sqrt{\lambda},1)} \sqrt{\frac{1}{2T}\log \frac{1}{Z}} \sum_{j=2}^{K} \sqrt{2^j} + \frac{D}{\max(\sqrt{\lambda},1)} \frac{Z(K-1)}{4}\\
\leq & \frac{D}{\lambda} + \frac{D}{\max(\sqrt{\lambda},1)} \sqrt{\frac{1}{2T}\log \frac{1}{Z}} \frac{2}{\sqrt{2}-1} \sqrt{2}^{K-1}+ \frac{D}{\max(\sqrt{\lambda},1)} \frac{Z(K-1)}{4}\\
\overset{(\ref{eqn:def:K})}{\leq}& \frac{D}{\lambda} + \frac{D}{\max(\sqrt{\lambda},1)} \sqrt{\frac{1}{2T}\log \frac{1}{Z}} \frac{2}{\sqrt{2}-1} \sqrt{\frac{T}{ 32 \lambda \log 1/Z}} + \frac{D}{\max(\sqrt{\lambda},1)} \frac{Z}{4} \log_2 T \\
= & \frac{D}{\lambda} + \frac{(\sqrt{2}+1)D}{4 \max(\sqrt{\lambda},1) \sqrt{\lambda}} + \frac{ D \log_2 T}{4 \max(\sqrt{\lambda},1)  T} \overset{(\ref{eqn:T:lower})}{\leq} \frac{2D}{\lambda}
\end{split}
\end{equation}
implies that $\bv_t^i$'s meet the condition in (\ref{eqn:switch:expert}) when $M=2$.

Since $\B^1=A^1$, we have
\begin{equation} \label{eqn:movement:B1}
\|\bv_t^1-\bv_{t+1}^1\| = \|\w_t^1-\w_{t+1}^1\| \overset{(\ref{eqn:movement:OGD})}{\leq}  \frac{D}{\lambda}.
\end{equation}
Thus, (\ref{eqn:movement:BiS}) holds when $i=1$. Suppose (\ref{eqn:movement:BiS}) is true when $i=k$, and thus
\begin{equation} \label{eqn:movement:Bk}
 \|\bv_t^k-\bv_{t+1}^k\| \leq \frac{D}{\lambda} + \frac{D}{\max(\sqrt{\lambda},1)}  \sum_{j=2}^{k} \left( \sqrt{\frac{1}{\leh^{(j)}} \log \frac{1}{Z}}+ \frac{Z}{4} \right)  \overset{(\ref{eqn:upper:movement})}{\leq} \frac{2D}{\lambda}.
\end{equation}
We proceed to bound the movement of $\bv_t^{k+1}$, which is the output of $B^{k+1}$. Recall that $B^{k+1}$ is an instance of Combiner which aggregates $\B^k$ and $\A^{k+1}$. From the procedure of Algorithm~\ref{alg:3}, we have
\[
\bv_t^{k+1} \overset{(\ref{eqn:weight:combin})}{=} (1-w_t^{k+1})  \bv_t^{k} + w_t^{k+1}  \w_t^{k+1}
\]
where $w_t^{k+1}$ is the weight generated by DNP-cu. Thus, the movement of $\bv_t^{k+1}$ can be bounded by
\begin{equation} \label{eqn:movement:vk1}
\begin{split}
& \|\bv_t^{k+1}-\bv_{t+1}^{k+1}\| = \left\|(1-w_t^{k+1})  \bv_t^k +  w_t^{k+1} \w_t^{k+1} -\big( (1-w_{t+1}^{k+1})\bv_{t+1}^k +  w_{t+1}^{k+1} \w_{t+1}^{k+1} \big) \right\| \\
\leq &\left\|(1-w_t^{k+1}) \bv_t^k  -(1-w_{t+1}^{k+1}) \bv_t^k + w_t^{k+1} \w_t^{k+1} -w_{t+1}^{k+1} \w_t^{k+1} \right\| \\
&+ \left\|(1-w_{t+1}^{k+1}) \bv_t^k + w_{t+1}^{k+1} \w_t^{k+1} -\big( (1-w_{t+1}^{k+1}) \bv_{t+1}^k + w_{t+1}^{k+1} \w_{t+1}^{k+1} \big) \right\|\\
\leq & |w_t^{k+1} -w_{t+1}^{k+1}| \|\bv_t^k-\w_t^{k+1}\| + (1-w_{t+1}^{k+1}) \|\bv_t^k-\bv_{t+1}^k\| + w_{t+1}^{k+1} \|\w_t^{k+1}-\w_{t+1}^{k+1}\| \\
\overset{(\ref{eqn:domain}), (\ref{eqn:movement:OGD})}{\leq} & D |w_t^{k+1} -w_{t+1}^{k+1}|  +  (1-w_{t+1}^{k+1})\|\bv_t^k-\bv_{t+1}^k\| + w_{t+1}^{k+1} \frac{D}{\lambda}\\
\leq & D |w_t^{k+1} -w_{t+1}^{k+1}|  + \max \left( \|\bv_t^k-\bv_{t+1}^k\|, \frac{D}{\lambda} \right).
\end{split}
\end{equation}

From (\ref{eqn:movement:OGD}) and (\ref{eqn:movement:Bk}), we know that the outputs of $\B^k$ and $\A^{k+1}$ satisfy (\ref{eqn:switch:expert}). Thus, we can apply Corollary~\ref{cor:1} to bound the change of $w_t^{k+1}$:
\begin{equation} \label{eqn:movement:wk1}
|w_t^{k+1} -w_{t+1}^{k+1}| \overset{(\ref{eqn:change:prediction:F:3})}{\leq} \frac{1}{\max(\sqrt{\lambda},1)}  \left( \sqrt{\frac{1}{\leh^{(k+1)}} \log \frac{1}{Z}}+ \frac{Z}{4} \right).
\end{equation}
From (\ref{eqn:movement:vk1}) and (\ref{eqn:movement:wk1}), we have
\[
\begin{split}
 \|\bv_t^{k+1}-\bv_{t+1}^{k+1}\| \leq &  \frac{D}{\max(\sqrt{\lambda},1)}  \left( \sqrt{\frac{1}{\leh^{(k+1)}} \log \frac{1}{Z}}+ \frac{Z}{4} \right)  + \max \left( \|\bv_t^k-\bv_{t+1}^k\|, \frac{D}{\lambda} \right) \\
\overset{(\ref{eqn:movement:Bk})}{\leq} & \frac{D}{\lambda} + \frac{D}{\max(\sqrt{\lambda},1)}  \sum_{j=2}^{k+1} \left( \sqrt{\frac{1}{\leh^{(j)}} \log \frac{1}{Z}}+ \frac{Z}{4} \right)
\end{split}
\]
which shows that (\ref{eqn:movement:BiS}) holds when $i=k+1$.

\subsection{Proof of Lemma~\ref{lem:BK:Bk}}
The regret of $\B^K$ w.r.t.~$\A^k$  can be decomposed as
\begin{equation} \label{eqn:decomp:bound}
\begin{split}
&\sum_{t=r}^s \big( f_t(\bv_t^K)+  \lambda G  \|\bv_t^K-\bv_{t+1}^K\|\big) - \sum_{t=r}^s \big( f_t(\w_t^k)+  \lambda G  \|\w_t^k-\w_{t+1}^k\|\big) \\
= &\underbrace{\sum_{t=r}^s \big( f_t(\bv_t^k)+  \lambda G  \|\bv_t^k-\bv_{t+1}^k\|\big) - \sum_{t=r}^s \big( f_t(\w_t^k)+  \lambda G  \|\w_t^k-\w_{t+1}^k\|\big)}_{:=U} \\
+& \sum_{i=k+1}^{K}  \left(\underbrace{\sum_{t=r}^s \big( f_t(\bv_t^{i})+  \lambda G  \|\bv_t^{i}-\bv_{t}^{i+1}\|\big) - \sum_{t=r}^s \big( f_t(\bv_t^{i-1})+  \lambda G  \|\bv_t^{i-1}-\bv_{t+1}^{i-1}\|\big)}_{:=V^{i}} \right)
\end{split}
\end{equation}
where $U$ is the regret of $\B^k$ w.r.t.~$\A^k$, and $V^{i}$ is the regret of $\B^{i}$ w.r.t.~$\B^{i-1}$. Next, we make use of Corollary~\ref{cor:1} and Lemma~\ref{lem:combine:two} to bound those quantities.

To bound $U$, we have
\begin{equation} \label{eqn:U:bound}
\begin{split}
  U \overset{(\ref{eqn:cor:lower}),(\ref{eqn:com:reg:2})}{\leq }  &3GD  \bigg(  \max(\sqrt{\lambda},1) U(\leh^{(k)}) \left(\frac{\tau}{\leh^{(k)}} +1 \right) + \frac{2\tau}{\leh^{(k)}}  +1  +\max(\sqrt{\lambda},1) U(\leh^{(k)}) \\
  &+1  +\max(\sqrt{\lambda},1) \frac{\tau}{T} \bigg) \\
\overset{\tau \leq c \leh^{(k)}}{\leq }  &3GD  \left( (2+c) \max(\sqrt{\lambda},1) U(\leh^{(k)})  +2+2c +\max(\sqrt{\lambda},1)  \right) \\
\overset{(\ref{eqn:u:tau})}{\leq } &  3GD  \left( (2+c) \max(\sqrt{\lambda},1)\sqrt{16\leh^{(k)} \log T}  +2+2c +\max(\sqrt{\lambda},1)  \right)\\
\leq &3GD \max(\sqrt{\lambda},1) \left( 4(2+c) \sqrt{\leh^{(k)} \log T}  +3+2c   \right).
\end{split}
\end{equation}

To bound the summation of $V^i$, we have
\begin{equation} \label{eqn:Vi:bound}
\begin{split}
\sum_{i=k+1}^{K} V^i \overset{(\ref{eqn:cor:lower}),(\ref{eqn:com:reg:1})}{\leq }  & 3 GD \sum_{i=k+1}^{K} \left( \max(\sqrt{\lambda},1) U(\leh^{(i)}) +1  +\max(\sqrt{\lambda},1) \frac{\tau}{T} \right) \\
\leq & 3 GD \max(\sqrt{\lambda},1) \sum_{i=k+1}^{K} U(\leh^{(i)}) +  6  GD  \max(\sqrt{\lambda},1)(K-k) \\
\overset{(\ref{eqn:u:tau})}{\leq } & 3 GD \max(\sqrt{\lambda},1) \sum_{i=k+1}^{K} \sqrt{16\leh^{(i)} \log T} +  6  GD  \max(\sqrt{\lambda},1)(K-k) \\
\overset{(\ref{eqn:tau:value})}{= }&12 GD \max(\sqrt{\lambda},1) \sqrt{\leh^{(k)} \log T  } \sum_{i=1}^{K-k} \sqrt{2^{-i} } +  6  GD  \max(\sqrt{\lambda},1) (K-k)\\
\leq & 12 GD \max(\sqrt{\lambda},1) \sqrt{\leh^{(k)} \log T} \frac{1}{\sqrt{2}-1} +  6  GD \max(\sqrt{\lambda},1) (K-k) \\
\leq & 29 GD \max(\sqrt{\lambda},1) \sqrt{\leh^{(k)} \log T}  +  6  GD \max(\sqrt{\lambda},1) (K-k).
\end{split}
\end{equation}

We complete the proof by substituting  (\ref{eqn:U:bound}) and (\ref{eqn:Vi:bound}) into (\ref{eqn:decomp:bound}).
\section{Conclusion and Future Work}
Based on a variant of Discounted-Normal-Predictor (DNP-cu), we design a novel algorithm, named as smoothed OGD for SOCO. Our algorithm combines multiple instances of OGD sequentially by DNP-cu, and thus is very simple. Theoretical analysis shows that it attains nearly optimal bounds for both adaptive regret and dynamic regret over any interval, in the presence of switching costs.

In the studies of adaptive regret and dynamic regret, we can obtain tighter bounds when the hitting cost exhibits additional curvature properties such as exponential concavity \citep{Adaptive:Hazan,pmlr-v134-baby21a} and smoothness  \citep{Adaptive:Regret:Smooth:ICML,Problem:Dynamic:Regret}. It remains unclear whether DNP-cu can exploit such information to further improve the performance. For SOCO, it is common to consider the lookahead setting, and the problem is still nontrivial due to the coupling created by the switching cost \citep{SOCO:OBD}. For dynamic regret with switching cost, \citet{Smoothed:Online:NeurIPS} have demonstrated that Assumption~\ref{ass:1} is unnecessary in the lookahead setting. It would be interesting to develop a lookahead version of DNP-cu, and verify whether we can drop Assumption~\ref{ass:1} as well.

\bibliography{E:/MyPaper/ref}
\appendix

\section{Proof of Theorem~\ref{thm:2}}\label{sec:proof:thm2}
Our purpose is to provide a general analysis of Algorithm~\ref{alg:1} over any bit sequence, so we do not make use of the range of $x_t$ in (\ref{eqn:xt:range}). As an alternative, we use the following simple upper bound
\begin{equation} \label{eqn:xt:bound}
|x_t| \leq \bb \leh , \ \forall  t \geq 1
\end{equation}
which can be proved by induction. From the initialization, we have $|x_1| =0 \leq \bb \leh$. Now, suppose $|x_k| \leq \bb \leh $. Then,  we have
\[
|x_{k+1}| \leq |\rho x_k|+|b_k| \overset{(\ref{eqn:discount})}{\leq} \left( 1-\frac{1}{\leh}\right) \bb \leh   + \bb = \bb \leh .
\]
Then, we can bound the difference between any two consecutive derivations by $2$:
\begin{equation} \label{eqn:xt:difference}
|x_t - x_{t+1}| = |(1-\rho) x_t  - b_t | \overset{(\ref{eqn:discount})}{\leq}   \frac{1}{\leh} |x_t| + |b_t| \overset{(\ref{eqn:xt:bound})}{\leq} 2 \bb \leq 2.
\end{equation}

Next, we introduce Lemma 19 of \citet{pmlr-v98-daniely19a}, which characterizes the derivative of $g(\cdot)$ over short intervals.
\begin{lemma}\label{lem:derivative} Suppose $\log \frac{1}{Z} \leq \frac{\leh}{16}$, $Z \leq \frac{1}{e}$ and $\leh \geq 8e$. For every segment $\I \subset \R$ of length $\leq 2$ and every $x \in \I$, we have
\[
4 \max_{s \in \I} |g'(s)| \leq \frac{1}{\leh} x g(x) +Z .
\]
\end{lemma}
Then, we can apply the above lemma to bound the derivative of $g(\cdot)$ over the interval $[x_t, x_{t+1}]$,\footnote{With a slight abuse of notation,  we will write $[a,b]$ to denote $[\min\{a,b\},\max\{a,b\}]$.} whose length is smaller than $2$. Under the conditions of Lemma~\ref{lem:derivative}, we have
\begin{equation} \label{eqn:grad:bound1}
4 \max_{s \in [x_t, x_{t+1}]} |g'(s)| \leq \frac{1}{\leh} x_t g(x_t) +Z.
\end{equation}
Since $g(x) =0$ if $x \leq 0$, we have
\begin{equation} \label{eqn:grad:bound2}
4 \max_{s \in [x_t, x_{t+1}]} |g'(s)| \leq  Z,  \textrm{ if } x_t \leq 0.
\end{equation}
Furthermore, we know that $g'(x)=0$, if $x \geq U(\leh)$. When $x_t \geq  U(\leh)+2 \bb$, from (\ref{eqn:xt:difference}) we have
\[
[x_t, x_{t+1}] \subset  [U(\leh), \infty )  .
\]
Thus,
\begin{equation} \label{eqn:grad:bound3}
\max_{s \in [x_t, x_{t+1}]} |g'(s)| =0, \textrm{ if } x_t  \geq U(\leh)+2 \bb.
\end{equation}
Let $\ind(x)$ be the indicator function of the interval $[0, U(\leh)+2 \bb]$. We can summarize the general result in (\ref{eqn:grad:bound1}) and the special cases in (\ref{eqn:grad:bound2}) and (\ref{eqn:grad:bound3}) as
\begin{equation} \label{eqn:grad:bound}
4 \max_{s \in [x_t, x_{t+1}]} |g'(s)| \leq \frac{1}{\leh} x_t g(x_t) \ind(x_t)+Z.
\end{equation}

We proceed to use the following  potential function
\[
\Phi_t = \int_0^{x_t} g(s) ds
\]
to analyze the reward of Algorithm~\ref{alg:1}.  It is easy to verify that
\begin{equation} \label{eqn:phi:upper}
\max(0, x_t-U(\leh)) \leq \Phi_t = \int_0^{x_t} g(s) ds \leq \max(x_t,0).
\end{equation}
To bound the change of the potential function, we need the following inequality for piece-wise differential functions $f:[a,b] \mapsto \R$ \citep{DNP:2010:Arxiv,pmlr-v98-daniely19a}
\begin{equation} \label{eqn:integral}
\int_a^b f(x) dx \leq  f(a)(b-a) + \max|f'(z)| \frac{1}{2} (b-a)^2 .
\end{equation}
We have
\begin{equation} \label{eqn:diff:poten}
\begin{split}
\Phi_{t+1} -\Phi_t = & \int_{x_t}^{x_{t+1}} g(s) ds \\
\overset{(\ref{eqn:integral})}{\leq} & g(x_t) ( x_{t+1}- x_{t}) + \frac{1}{2} ( x_{t+1}- x_{t})^2 \max_{s\in [x_t, x_{t+1}]} |g'(s)| \\
\overset{(\ref{eqn:xt:difference})}{\leq} & g(x_t) \left( -\frac{1}{\leh} x_t + b_t \right) + 2  \max_{s\in [x_t, x_{t+1}]} |g'(s)| \\
=& g(x_t) \left( -\frac{1}{\leh} x_t + b_t \right) - 2  \max_{s\in [x_t, x_{t+1}]} |g'(s)|  + 4 \max_{s\in [x_t, x_{t+1}]} |g'(s)| \\
\leq & g(x_t) \left( -\frac{1}{\leh} x_t + b_t \right) - 2  \left| \frac{g(x_t)-g(x_{t+1})}{x_t -x_{t+1}} \right|  + 4 \max_{s\in [x_t, x_{t+1}]} |g'(s)| \\
\overset{(\ref{eqn:xt:difference})}{\leq}& g(x_t) \left( -\frac{1}{\leh} x_t + b_t \right) - \frac{1}{\bb} |g(x_t)-g(x_{t+1})| + 4 \max_{s\in [x_t, x_{t+1}]} |g'(s)| \\
\overset{(\ref{eqn:grad:bound})}{\leq}& g(x_t) \left( -\frac{1}{\leh} x_t + b_t \right) - \frac{1}{\bb} |g(x_t)-g(x_{t+1})| +  \frac{1}{\leh} x_t g(x_t) \ind(x_t)+Z  \\
=&  g(x_t) b_t  -   \frac{1}{\bb} |g(x_t)-g(x_{t+1})| +   \frac{1}{\leh} x_t g(x_t) \left( \ind(x_t)   -1 \right)+ Z
\end{split}
\end{equation}
where the 3rd inequality is due to the mean value theorem.  To bound the cumulative reward over any interval $[r,s]$, we sum (\ref{eqn:diff:poten}) from $t=r$ to $t=s$, and obtain
\[
\begin{split}
 \Phi_{s+1} -\Phi_r \leq   \sum_{t=r}^s \left( g(x_t) b_t -   \frac{1}{\bb} |g(x_t)-g(x_{t+1})| \right) + \sum_{t=r}^s \frac{1}{\leh} x_t g(x_t) \left( \ind(x_t)   -1 \right)+ Z \tau.
\end{split}
\]
Thus,
\begin{equation} \label{eqn:sum:poten}
\begin{split}
\sum_{t=r}^s \left( g(x_t) b_t -   \frac{1}{\bb} |g(x_t)-g(x_{t+1})| \right) \geq  \Phi_{s+1}  + \sum_{t=r}^s \frac{1}{\leh} x_t g(x_t) \left( 1-\ind(x_t)   \right) -\Phi_r - Z \tau.
\end{split}
\end{equation}

First, we use the simple fact that
\[
 x_t g(x_t) \left( 1 -\ind(x_t)  \right) \geq  0,
\]
to simplify (\ref{eqn:sum:poten}), and have
\begin{equation} \label{eqn:lower:bound:1}
\begin{split}
\sum_{t=r}^s \left( g(x_t) b_t -   \frac{1}{\bb} |g(x_t)-g(x_{t+1})| \right) \geq  -\Phi_r - Z \tau.
\end{split}
\end{equation}

Second, we lower bound the cumulative reward by the summation of the bit sequence. From (\ref{eqn:phi:upper}), we have
\begin{equation} \label{eqn:phi:lower}
\Phi_{s+1}   \geq x_{s+1} - U (\leh).
\end{equation}
We also have
\begin{equation} \label{eqn:product:lower}
x_t g(x_t) \left(  1- \ind(x_t) \right) \geq x_t - U(\leh) -2 \mu.
\end{equation}
That is because if $x_t \geq U(\leh) +2 \mu $, we have
\[
x_t g(x_t) \left(  1- \ind(x_t) \right)  =x_t \geq x_t - U(\leh) -2 \mu;
\]
otherwise,
\[
x_t g(x_t) \left(  1- \ind(x_t) \right) \geq 0 \geq x_t - U(\leh) -2 \mu.
\]
Based on (\ref{eqn:phi:lower}) and (\ref{eqn:product:lower}), we have
\[
\begin{split}
&\Phi_{s+1} + \sum_{t=r}^{s} \frac{1}{\leh} x_t g(x_t) \left(1- \ind(x_t)  \right) \\
\geq & x_{s+1} - U(\leh)   + \frac{1}{\leh} \sum_{t=r}^{s}   \left(x_t - U(\leh) -2 \mu \right)  = x_{s+1} + \frac{1}{\leh} \sum_{t=r}^{s}  x_t - \frac{\tau}{\leh} \left( U(\leh) +2 \mu\right) - U(\leh) \\
=& \rho^{\tau} x_r + \sum_{j=r}^{s} \rho^{s-j} b_j + \frac{1}{\leh} \sum_{t=r}^{s}  \left(\rho^{t-r} x_r +  \sum_{j=r}^{t-1} \rho^{t-1-j} b_j \right)  - \frac{\tau}{\leh} \left( U(\leh) +2 \mu\right) - U(\leh)\\
=& \rho^{\tau} x_r +  \sum_{j=r}^{s} \rho^{s-j} b_j +  \frac{x_r}{\leh}   \sum_{t=r}^{s}  \rho^{t-r} +     \sum_{j=r}^{s}  \frac{b_j}{\leh} \sum_{t=j+1}^{s}   \rho^{t-1-j}   - \frac{\tau}{\leh} \left( U(\leh) +2 \mu\right) - U(\leh)\\
=& \rho^{\tau} x_r +  \sum_{j=r}^{s} \rho^{s-j} b_j +  \frac{x_r}{\leh}    \frac{1-\rho^{\tau}}{1-\rho}+     \sum_{j=r}^{s}  \frac{b_j}{\leh} \frac{1-\rho^{s-j}}{1-\rho}   - \frac{\tau}{\leh} \left( U(\leh) +2 \mu\right) - U(\leh)\\
\overset{(\ref{eqn:discount})}{=}&x_r + \sum_{t=r}^s b_t  - \frac{\tau}{\leh} \left( U(\leh) +2 \mu\right) - U(\leh).
\end{split}
\]
Combining the above inequality with (\ref{eqn:sum:poten}), we have
\begin{equation} \label{eqn:lower:bound:2}
\begin{split}
\sum_{t=r}^s \left( g(x_t) b_t -   \frac{1}{\bb} |g(x_t)-g(x_{t+1})| \right)
 \geq  \sum_{t=r}^s b_t + x_r  - \frac{\tau}{\leh} \left( U(\leh) +2 \mu\right) - U(\leh) -\Phi_r - Z \tau.
\end{split}
\end{equation}

Third, from (\ref{eqn:lower:bound:1}) and (\ref{eqn:lower:bound:2}), we have
\[
\begin{split}
&\sum_{t=r}^s \left( g(x_t) b_t -   \frac{1}{\bb} |g(x_t)-g(x_{t+1})| \right)\\
\geq  &\max\left(0, \sum_{t=r}^s b_t + x_r  - \frac{\tau}{\leh} \left( U(\leh) +2 \mu\right) - U(\leh) \right)  -\Phi_r - Z \tau\\
\overset{(\ref{eqn:phi:upper})}{\geq}  &\max\left(0, \sum_{t=r}^s b_t + x_r  - \frac{\tau}{\leh} \left( U(\leh) +2 \mu\right) - U(\leh) \right)  -  \max(x_r,0)- Z \tau
\end{split}
\]
which proves (\ref{eqn:alg1:lower}).

Finally, to bound the change of successive predictions, we have
\begin{equation} \label{eqn:change:prediction}
|g(x_t)-g(x_{t+1})| \leq |x_t -x_{t+1}| \max_{s} |g'(s)| \overset{(\ref{eqn:xt:difference})}{\leq} 2 \bb \max_{s} |g'(s)|.
\end{equation}
Following the analysis of Lemma 23 of \citet{pmlr-v98-daniely19a}, we know that $g'(\cdot)$ is nondecreasing in $[0, U(\leh)]$ and is $0$ outside, and thus
\begin{equation} \label{eqn:max:derivative}
\max_{s} |g'(s)|= g'(U(\leh)) = \frac{U(\leh) g(U(\leh))}{8 \leh} + \frac{Z}{8} = \frac{U(\leh) }{8 \leh} + \frac{Z}{8} \overset{(\ref{eqn:u:tau})}{\leq} \frac{\sqrt{16\leh \log \frac{1}{Z}}}{8 \leh} + \frac{Z}{8}
\end{equation}
where the 2nd equality is due to the property of the confidence function \citep[Lemma 18]{pmlr-v98-daniely19a}. We obtain (\ref{eqn:change:prediction:F}) by combining (\ref{eqn:change:prediction}) and (\ref{eqn:max:derivative}).

\end{document}